\documentclass[12pt,a4paper]{article}

\usepackage[T1]{fontenc}
\usepackage[utf8]{inputenc}

\usepackage[a4paper,margin=1in]{geometry}
\usepackage{setspace}
\usepackage{amsmath,amssymb}
\usepackage{graphicx}

\usepackage{booktabs}
\usepackage{tabularx}
\usepackage{array}
\usepackage{longtable}
\usepackage{multirow}
\usepackage{pdflscape}

\usepackage{caption}
\usepackage{subcaption}
\usepackage{float}
\usepackage{placeins}

\usepackage{natbib}

\usepackage[expansion=false]{microtype}
\usepackage{fancyhdr}
\usepackage[hidelinks]{hyperref}

\renewcommand{\arraystretch}{1.15}

\title{Benchmarking ConvLSTM for One-Day-Ahead IMDAA Rainfall-Field
Prediction across Four Indian Cities}

\author{
Tanmay Ghosh\thanks{
National Institute of Advanced Studies, Indian Institute of Science Campus,
Bengaluru, India.
\texttt{tanmayghosh10@gmail.com}
(Corresponding author)
}
\and
Shaurabh Anand\thanks{
School of Development, Azim Premji University, Bengaluru, India.
\texttt{shaurabh.anand@apu.edu.in}
}
\and
Rakesh Gomaji Nannewar\thanks{
National Institute of Advanced Studies, Indian Institute of Science Campus,
Bengaluru, India.
\texttt{rakeshn@nias.res.in}
}
\and
Nithin Nagaraj\thanks{
Complex Systems Programme, National Institute of Advanced Studies,
Indian Institute of Science Campus, Bengaluru, India.
\texttt{nithin@nias.res.in}
}
}

\date{}

\begin{document}

\maketitle

\begin{abstract}

Convolutional long short-term memory networks (ConvLSTMs) are widely used for
precipitation forecasting, but most evidence for their performance comes from
dense, high-frequency radar sequences. This study tests whether convolutional
recurrence improves one-day-ahead rainfall-field prediction on small daily
reanalysis grids. Indian Monsoon Data Assimilation and Analysis (IMDAA) fields
for June--September 1998--2020 were analysed for Bengaluru, Delhi, Kolkata and
Mumbai. Ten na\"ive, statistical, tree-based and neural approaches were
compared using atmospheric-only and rainfall-history-plus-atmospheric inputs.
Performance was assessed for complete fields, domain-mean rainfall, spatial
anomalies and high-rainfall days.

ConvLSTM did not consistently outperform simpler alternatives. FC-LSTM
produced the numerically lowest domain-mean rainfall error in Bengaluru,
Kolkata and Mumbai, whereas persistence performed best in Delhi. ConvLSTM
produced the numerically lowest spatial-anomaly error only in Mumbai, where
rainfall fields also showed greater short-term spatial continuity and rainfall
history improved all three neural architectures. Its difference from FC-LSTM
was nevertheless small. Neural models underestimated rainfall magnitude and
detected too few threshold exceedances on high-rainfall days, while persistence
achieved the highest detection performance in every city. Post-hoc analyses
showed that the selected models were most sensitive to the latest input day,
with broader recent-lag sensitivity in Mumbai. The findings show that gridded
inputs alone do not justify ConvLSTM and that architecture choice should follow
strong benchmarking across average, spatial and high-rainfall performance.

\end{abstract}



\section{Introduction}
\label{sec:introduction}

Rainfall over Indian cities varies substantially from one day to the next and
across city-scale domains. Short-range rainfall-field prediction must therefore
represent both the average rainfall over a domain and its spatial distribution.
These properties are not necessarily reproduced equally well by forecasting
models. A model may estimate domain-mean rainfall accurately while smoothing
differences among grid cells, or it may preserve spatial variation while
remaining biased in the overall rainfall amount. Performance on high-rainfall
days also requires separate examination because these days occur less
frequently than dry and moderate conditions, and good overall performance does
not necessarily imply accurate prediction of larger rainfall amounts.
Rainfall-field prediction should therefore be evaluated in terms of rainfall
amount, spatial distribution and performance on high-rainfall days rather than
through a single summary measure.

Long short-term memory networks (LSTMs) are designed to learn dependencies in
sequential data, while convolutional neural networks (CNNs) preserve local
spatial relationships in gridded inputs. The convolutional long short-term
memory network (ConvLSTM) combines these functions by replacing the fully
connected transitions of a conventional LSTM with convolutional operations
\citep{Shi2015}. ConvLSTM was initially developed for precipitation nowcasting,
where consecutive radar images show the movement and development of rainfall
systems over large, densely sampled domains. Subsequent radar-based studies
showed that convolutional recurrent models can represent spatial and temporal
changes in precipitation, but they also identified important limitations.
Deterministic neural forecasts may become increasingly smooth and lose accuracy
at higher rainfall intensities \citep{Shi2017,Ayzel2020,Ravuri2021}.
Intensity-sensitive loss functions, generative models and additional
atmospheric predictors have consequently been investigated to improve the
representation of heavier rainfall
\citep{CambierVanNooten2023,Kim2024,Zhang2023}.

Daily reanalysis fields present a different prediction setting from radar
nowcasting. Radar datasets commonly contain observations every few minutes over
hundreds or thousands of spatial cells, allowing the movement and development
of rainfall systems to be observed directly. Daily reanalysis combines this
sub-daily evolution into a single field, while a city-centred domain may contain
only a small number of grid cells. In the present study, the four city domains
contain between 9 and 42 cells at a spacing of approximately
$0.12^{\circ}$. A fully connected long short-term memory network (FC-LSTM)
processes the complete field at each time step after flattening it, whereas
ConvLSTM applies local convolutional operations within its recurrent states.
These operations may be useful when successive rainfall fields retain coherent
spatial patterns, but they may contribute little when day-to-day spatial
continuity is weak. Previous studies have also shown that model rankings differ
across locations, input combinations and evaluation measures
\citep{Kumar2023,Pawar2024,Panda2024}. The relative performance of ConvLSTM on
small daily reanalysis grids is therefore evaluated through comparison with
reference forecasts, statistical and tree-based models, and neural
architectures that represent spatial and temporal information differently.

The information supplied to a model must also be distinguished from the
architecture used to process it. Previous rainfall fields provide direct
information about persistence and the continuation of recent spatial patterns.
Atmospheric variables such as relative humidity, cloud cover, pressure,
evaporation and wind provide information about the meteorological conditions
associated with subsequent rainfall. If prediction improves after previous
rainfall is included, the improvement may result from the additional rainfall
information, from the model's use of that information or from an interaction
between the input and the architecture. It cannot be attributed to architecture
alone. The present study therefore compares an atmospheric-only input
formulation with a second formulation that adds previous rainfall fields while
retaining the same atmospheric predictors, prediction target and model
structures.

Forecast performance must be evaluated against credible reference predictions.
Persistence assumes that the following rainfall field will resemble the latest
field, whereas day-of-year climatology represents the expected seasonal field
at each grid cell. A fitted model may improve on one reference but not the
other, so the interpretation of forecast skill depends on the benchmark used
\citep{Murphy1992}. The present study reports performance relative to both
persistence and climatology and uses the lower-RMSE na\"ive reference for each
city and evaluation metric when summarising forecast skill.

Evaluation must also distinguish among different properties of a predicted
rainfall field. Complete-field root-mean-square error (RMSE) combines errors in
rainfall amount and spatial distribution. Domain-mean rainfall RMSE evaluates
the average rainfall over the retained grid, whereas spatial-anomaly RMSE
evaluates within-field variation after the daily domain mean has been removed.
A model may perform well for one property and poorly for another, and no single
measure captures every relevant characteristic of a rainfall field
\citep{RobertsLean2008,Ebert2008}. Multi-scale neighbourhood verification is
valuable for large precipitation fields, but the grids used here contain too
few cells to support a meaningful assessment across several spatial scales.
Complete-field RMSE, domain-mean rainfall RMSE and spatial-anomaly RMSE are
therefore reported separately.

Average error may conceal poor performance on high-rainfall days. Dry, light
and moderate rainfall days form most of the sample, while errors associated
with larger rainfall amounts may differ systematically from those observed
under more common conditions. A model can therefore achieve a relatively low
overall error while underestimating larger rainfall amounts or predicting too
few threshold exceedances. High-rainfall performance is evaluated separately
using continuous measures of error and bias together with categorical detection
measures \citep{Stephenson2008,FerroStephenson2011}. High-rainfall days are
defined using the city-specific 90th percentile of training-period domain-mean
rainfall. They are described as high-rainfall days rather than meteorological
extremes because the threshold is derived from the study data rather than from
an externally defined extreme-rainfall standard.

The analysis uses the Indian Monsoon Data Assimilation and Analysis (IMDAA)
regional reanalysis. IMDAA combines observations with a numerical
weather-prediction and data-assimilation system to produce spatially complete
meteorological fields over the Indian monsoon region \citep{Rani2021}. Its
consistent spatial and temporal coverage permits the same experiment to be
conducted for Bengaluru, Delhi, Kolkata and Mumbai. However, IMDAA
precipitation is a reanalysis product rather than an independent observation of
surface rainfall. The study therefore evaluates one-day-ahead hindcasting of
rainfall as represented within IMDAA. It does not evaluate the hindcasts
against independent rain-gauge or radar observations, and the spatial
resolution of the retained fields does not support street- or ward-level
interpretation.

Post-hoc explanation methods are applied after predictive performance has been
established. Grouped permutation and temporal and spatial occlusion are used to
examine model sensitivity to predictor groups, input days and grid cells, while
gradient-weighted class activation mapping (Grad-CAM) provides an exploratory
representation of positive activation within the selected ConvLSTM models
\citep{Selvaraju2017}. These methods describe the behaviour of fitted models
rather than the atmospheric causes of rainfall. Attribution results can vary
with model parameters, input perturbations and the explanation method used
\citep{Adebayo2018,Ghorbani2019,Ismail2020,Bommer2024}. Their results are
therefore interpreted as measures of fitted-model sensitivity, not as evidence
of physical causation \citep{OLoughlin2025}.

This study evaluates one-day-ahead monsoon rainfall-field hindcasting for
Bengaluru, Delhi, Kolkata and Mumbai using IMDAA regional reanalysis. Separate
city-specific models use seven consecutive daily fields to predict rainfall on
the following day. ConvLSTM is compared with reference forecasts, statistical
and tree-based models, a CNN without recurrence and an FC-LSTM. The analysis
addresses four questions: whether the fitted models improve on credible
reference forecasts and simpler alternatives; whether previous rainfall adds
predictive information beyond the atmospheric variables; how the relative
performance of ConvLSTM differs across cities and whether the observed
numerical patterns are consistent with differences in day-to-day spatial
continuity; and whether conclusions based on overall test-period performance
also hold on high-rainfall days. Post-hoc explanation methods are then used to
examine the sensitivity of the selected neural models to predictor groups,
input days and grid cells. The contribution of the study is a controlled
assessment of whether the additional spatial recurrence represented by
ConvLSTM lowers prediction error relative to simpler alternatives on small
daily IMDAA rainfall fields.


\section{Materials and methods}
\label{sec:materials_methods}


\subsection{Study domains and IMDAA data}
\label{subsec:study_domains_imdaa}

The study covers Bengaluru, Delhi, Kolkata and Mumbai. These four
geographically separated cities provide contrasting monsoon rainfall settings
for examining the contribution of rainfall history and convolutional
recurrence across city-specific domains. The cities were modelled independently;
observations from different cities were not pooled during model fitting.

Meteorological fields were obtained from the Indian Monsoon Data Assimilation
and Analysis (IMDAA) regional reanalysis through the NCMRWF Reanalysis Data
Service \citep{Rani2021}. Data were retained for the June--September monsoon
seasons from 1998 to 2020, giving 23 monsoon seasons. Four six-hourly records
were available for each day and were aggregated separately for every variable
and grid cell to create the daily modelling fields.

\begin{figure}[H]
    \centering
    \includegraphics[
        width=\textwidth,
        height=0.78\textheight,
        keepaspectratio
    ]{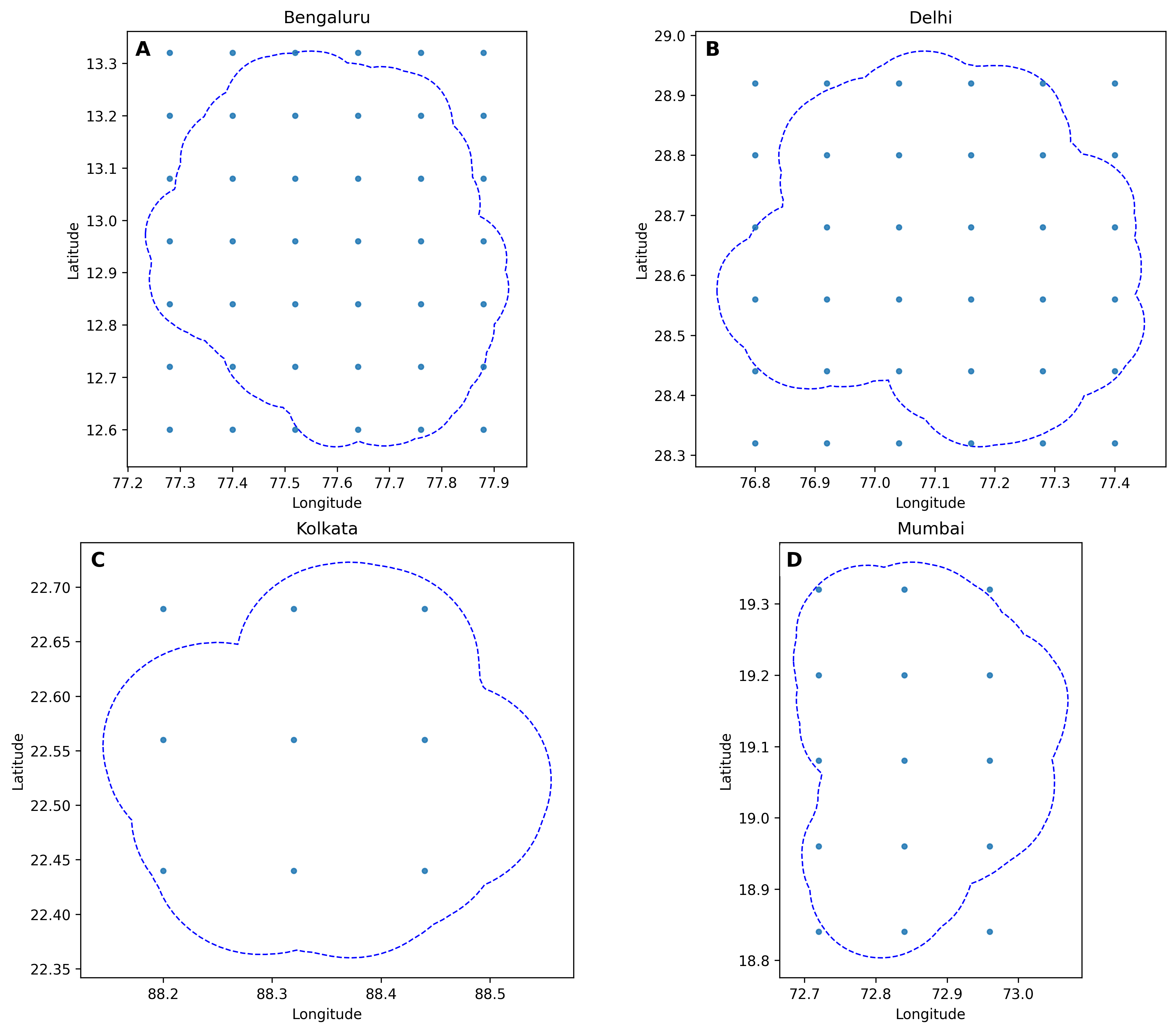}
    \caption{City-specific modelling domains and retained grid-point centres.
    Dashed outer boundaries are included for cartographic reference only and
    were not supplied to the forecasting models. All cells in each rectangular
    coordinate domain were retained.}
    \label{fig:study_area_grid}
\end{figure}

Adjacent grid-point centres in the prepared city extracts were separated by
$0.12^{\circ}$ in both latitude and longitude, corresponding to an approximate
spacing of 12~km. For each city, the modelling domain comprised the complete
rectangular set of unique latitude and longitude coordinates in the prepared
extract. Every cell within this rectangle was retained. Administrative
boundaries were used only as cartographic overlays and were neither used to
mask grid cells nor supplied as model predictors. The city extracts therefore
had different spatial dimensions, and no padding was applied to create a common
grid size. Figure~\ref{fig:study_area_grid} shows the retained grid-point
centres and the outer administrative boundaries used for cartographic
reference.

Table~\ref{tab:city_grids_samples} reports the coordinate ranges, spatial
dimensions and chronological sample sizes.

\begin{table}[H]
    \centering
    \scriptsize
    \setlength{\tabcolsep}{2.7pt}
    \caption{Prepared-grid coordinates, spatial dimensions and chronological
    sample sizes.}
    \label{tab:city_grids_samples}
    \begin{tabular}{@{}lccccccc@{}}
        \toprule
        \textbf{City} &
        \textbf{Latitude range} &
        \textbf{Longitude range} &
        \textbf{Grid} &
        \textbf{Cells} &
        \textbf{Training} &
        \textbf{Validation} &
        \textbf{Test} \\
        \midrule
        Bengaluru & $12.60$--$13.32$ & $77.28$--$77.88$ & $7\times6$ & 42 & 1,955 & 345 & 345 \\
        Delhi      & $28.32$--$28.92$ & $76.80$--$77.40$ & $6\times6$ & 36 & 1,955 & 345 & 345 \\
        Kolkata    & $22.44$--$22.68$ & $88.20$--$88.44$ & $3\times3$ & 9  & 1,955 & 345 & 345 \\
        Mumbai     & $18.84$--$19.32$ & $72.72$--$72.96$ & $5\times3$ & 15 & 1,955 & 345 & 345 \\
        \bottomrule
    \end{tabular}
    \begin{minipage}{0.96\textwidth}
        \vspace{3pt}\footnotesize
        \textit{Note:} Coordinates are grid-point centres in decimal degrees;
        adjacent points are separated by $0.12^{\circ}$. Training covers
        1998--2014, validation 2015--2017 and testing 2018--2020. Each sample
        contains seven consecutive input days and one following-day
        rainfall-field target.
    \end{minipage}
\end{table}

For total precipitation, the four six-hourly records were first averaged to
obtain the daily mean hourly rate. The resulting rate was multiplied by 24 to
express precipitation in mm day$^{-1}$:

\begin{equation}
    P_{d}
    =
    24
    \left(
        \frac{1}{4}
        \sum_{k=1}^{4} P_{d,k}
    \right),
    \label{eq:daily_precipitation}
\end{equation}

where $P_{d,k}$ is the precipitation rate for the $k$th six-hourly record on
day $d$. Negative precipitation values were set to zero before the logarithmic
transformations used in model fitting.

The prepared city-level datasets were checked for complete coverage across
dates, grid cells and the common predictor variables. A retained daily field
was required to contain all $H\times W$ grid cells.

The prediction target was the complete IMDAA total-precipitation field on the
day following the input sequence. The study therefore evaluates hindcasting of
precipitation as represented within IMDAA rather than accuracy against
independent rain-gauge or radar observations.


\subsection{Predictors and input formulations}
\label{subsec:predictors_inputs}

The same seven atmospheric predictors were used for all four cities:
evaporation rate, relative humidity, surface pressure, total cloud cover, zonal
wind, meridional wind and wind speed. These variables were used as supplied in
the prepared IMDAA datasets. Relative humidity was not derived from temperature
and dew-point temperature, and wind speed was not calculated from the zonal and
meridional wind components. Total cloud cover retained its supplied fractional
scale, while evaporation retained its supplied unit and sign convention. After
daily aggregation, the atmospheric predictors were transformed only by the
RobustScaler fitted to the training samples.

Convective rain rate and large-scale rain rate were excluded because they are
precipitation-related diagnostic variables closely connected to the prediction
target. Mean sea-level pressure, surface temperature, skin temperature and wind
gust were excluded because they were not consistently available across all four
prepared city datasets. Using the same atmospheric predictors in every city
prevented differences in model performance from being produced by different
predictor sets.

Two static grid-position channels were added to each daily field. These
represented the north--south and east--west position of each grid cell within
its city-specific tensor. Each channel was independently normalised to the
interval $[-1,1]$, remained constant through time and contained grid-position
indices rather than latitude and longitude measured in degrees.

Two input formulations were evaluated. The atmospheric-only formulation,
\texttt{ATM\_ONLY}, contained the seven atmospheric predictors and two
position channels, giving nine input channels. The
rainfall-history-plus-atmospheric formulation, \texttt{RAIN\_HIST\_ATM}, added
previous total precipitation at every input time, giving ten channels. Previous
precipitation was represented as $\log(1+P)$, where $P$ is daily
precipitation in mm day$^{-1}$. Comparing the two
formulations shows how prediction changed when previous rainfall was added
while retaining the same atmospheric predictors, position channels, target and
model architectures.

Table~\ref{tab:input_variables} reports the variables, modelling units, daily
treatment and inclusion in the two input formulations.

\begin{table*}[t]
    \centering
    \scriptsize
    \setlength{\tabcolsep}{2.0pt}
    \renewcommand{\arraystretch}{1.08}

    \caption{Variables, modelling units and treatment in the two input
    formulations.}
    \label{tab:input_variables}

    \begin{tabularx}{\textwidth}{
        >{\raggedright\arraybackslash}p{2.20cm}
        >{\raggedright\arraybackslash}p{1.80cm}
        >{\centering\arraybackslash}p{1.20cm}
        >{\raggedright\arraybackslash}X
        >{\centering\arraybackslash}p{1.10cm}
        >{\centering\arraybackslash}p{1.45cm}
        >{\raggedright\arraybackslash}p{1.55cm}}
        \toprule

        \textbf{Variable} &
        \textbf{Role} &
        \textbf{Unit} &
        \textbf{Daily aggregation and modelling treatment} &
        \textbf{ATM only} &
        \textbf{Rain + ATM} &
        \textbf{XAI group} \\

        \midrule

        Evaporation rate
        & Atmospheric predictor
        & mm\,day$^{-1}$
        & Mean of four six-hourly records; supplied sign retained; scaled
        using training data
        & Yes & Yes
        & Evaporation \\

        Relative humidity
        & Atmospheric predictor
        & \%
        & Mean of four six-hourly records; used as supplied; scaled using
        training data
        & Yes & Yes
        & Relative humidity \\

        Surface pressure
        & Atmospheric predictor
        & Pa
        & Mean of four six-hourly records; scaled using training data
        & Yes & Yes
        & Surface pressure \\

        Total cloud cover
        & Atmospheric predictor
        & Fraction
        & Mean of four six-hourly records; supplied fractional scale retained;
        scaled using training data
        & Yes & Yes
        & Cloud cover \\

        Zonal wind
        & Atmospheric predictor
        & m\,s$^{-1}$
        & Mean of four six-hourly records; scaled using training data
        & Yes & Yes
        & Wind \\

        Meridional wind
        & Atmospheric predictor
        & m\,s$^{-1}$
        & Mean of four six-hourly records; scaled using training data
        & Yes & Yes
        & Wind \\

        Wind speed
        & Atmospheric predictor
        & m\,s$^{-1}$
        & Mean of four six-hourly records; used as supplied; scaled using
        training data
        & Yes & Yes
        & Wind \\

        Previous total precipitation
        & Rainfall-history predictor
        & mm\,day$^{-1}$
        & Aggregated using Equation~\ref{eq:daily_precipitation}, followed by
        $\log(1+P)$
        & No & Yes
        & Past rainfall \\

        North--south grid index
        & Static position channel
        & ---
        & Normalised to $[-1,1]$ and repeated across all input days
        & Yes & Yes
        & Excluded \\

        East--west grid index
        & Static position channel
        & ---
        & Normalised to $[-1,1]$ and repeated across all input days
        & Yes & Yes
        & Excluded \\

        Next-day total precipitation
        & Prediction target
        & mm\,day$^{-1}$
        & Aggregated using Equation~\ref{eq:daily_precipitation}; fitted as
        $\log(1+P)$ and inverse-transformed before evaluation
        & Target & Target
        & Not applicable \\

        \bottomrule
    \end{tabularx}

    \begin{minipage}{0.98\textwidth}
        \footnotesize
        \textit{Note:} The same seven atmospheric predictors were used for
        Bengaluru, Delhi, Kolkata and Mumbai. An em dash in the unit column
        denotes a dimensionless variable. Atmospheric variables were
        transformed using a RobustScaler fitted to the training data. The
        rainfall-history and grid-position channels were not passed through
        this scaler. Grid-position channels were included as model inputs but
        excluded from the meteorological predictor-group interpretation in the
        post-hoc XAI analysis.
    \end{minipage}
\end{table*}

Prepared predictor-channel names, temporal alignment and predictor-specific
leakage safeguards are reported in Supplementary Table~S3.


\subsection{Sequence construction and chronological division}
\label{subsec:sequence_split}

For each city and input formulation, a daily input was represented as an
$H\times W\times F$ tensor and the corresponding precipitation field as an
$H\times W$ array. A forecasting sample combined seven consecutive input
fields from $t-6$ through $t$, with the following day's precipitation field as
the target:

\begin{equation}
    \mathbf{S}_{t}
    =
    \left[
        \mathbf{X}_{t-6},\ldots,\mathbf{X}_{t}
    \right]
    \in
    \mathbb{R}^{7\times H\times W\times F},
    \qquad
    \mathbf{Y}_{t}
    =
    \mathbf{P}_{t+1}
    \in
    \mathbb{R}^{H\times W\times1}.
    \label{eq:forecasting_sequence}
\end{equation}

A sequence was retained only when all seven input dates and the target date
were consecutive. The first input date and target date were also required to
occur within the same calendar year. Sequence generation was therefore reset
within every June--September season, and no sequence connected September of
one year to June of the following year.

Each 122-day June--September period produced 115 valid sequences. Models were
trained on target years 1998--2014, validated on 2015--2017 and tested on
2018--2020. This produced 1,955 training sequences, 345 validation sequences
and 345 test sequences for every city and input formulation.

The seven atmospheric channels were transformed using a \texttt{RobustScaler}
fitted to the training data. Each atmospheric predictor was centred using its
training-period median and divided by its training-period interquartile range.
The fitted transformation was applied unchanged to the validation and test
data. The rainfall-history channel was not passed through the atmospheric
scaler because it had already been transformed using $\log(1+P)$. The
normalised position channels were also left unchanged.

The same sequence-construction rules, year-based split and training-only
atmospheric scaling were used in the model-training and XAI stages. A dated
example of the seven-day input sequence, next-day target and leakage-control
steps is provided in Supplementary Figure~S6.


\subsection{Forecasting models}
\label{subsec:forecasting_models}

Ten forecasting approaches were compared. Five simple reference methods were
included: zero rainfall, training-period day-of-year climatology, persistence,
a three-day moving average and a seven-day moving average. Each reference
method produced a complete $H\times W$ rainfall field. Persistence used the
rainfall field on day $t$, while the moving-average forecasts used the mean of
the preceding three or seven rainfall fields. The climatological forecast was
calculated separately for each grid cell and target day of year using
training-period rainfall only.

Ridge regression and ExtraTrees were used as non-neural comparisons. For both
models, the seven-day input tensor was flattened into one predictor vector and
the next-day rainfall field was flattened into an $HW$-dimensional target.
Ridge regression used an $L_{2}$ penalty of $\alpha=1$. ExtraTrees used 400
trees, a minimum leaf size of 2 and square-root feature sampling.

Three neural architectures were evaluated. The CNN-only model stacked the
seven input days along the channel dimension and applied three
$3\times3$ convolutional layers before producing the rainfall field. It
represented spatial processing without recurrent state transitions.

The FC-LSTM model flattened each daily field and processed the resulting
seven-step sequence using an LSTM with 96 units, followed by a dense layer.
It represented temporal recurrence without preserving local neighbourhoods
within the recurrent transition.

The TimeDistributed-CNN--ConvLSTM model first applied a $3\times3$
convolutional encoder to every input day. The encoded sequence then passed
through two ConvLSTM layers with 24 and 16 filters, respectively. Two
subsequent convolutional layers refined the final spatial representation
before a $1\times1$ convolution produced the next-day rainfall field. The
architecture therefore combined daily spatial encoding, convolutional
recurrence and spatial output refinement.

The purpose of the neural comparison was not to introduce a new architecture.
CNN-only tested the value of spatial convolution without recurrence; FC-LSTM
tested recurrence over the complete field without local convolutional
transitions; and the TimeDistributed-CNN--ConvLSTM tested their combination.

Table~\ref{tab:model_comparison} summarises the ten forecasting approaches.
Layer-by-layer neural-network specifications are reported in Supplementary
Table~S4a, city- and formulation-specific tensor dimensions and trainable
parameter counts in Supplementary Table~S4b, and training, model-selection
and recorded software settings in Supplementary Table~S4c.

\begin{table}[H]
    \centering
    \footnotesize
    \setlength{\tabcolsep}{4pt}
    \caption{Models included in the forecasting comparison.}
    \label{tab:model_comparison}

    \begin{tabularx}{\textwidth}{
        >{\raggedright\arraybackslash}p{3.0cm}
        >{\raggedright\arraybackslash}p{3.3cm}
        >{\raggedright\arraybackslash}p{3.2cm}
        >{\raggedright\arraybackslash}X
    }

        \toprule

        \textbf{Model}
        &
        \textbf{Spatial treatment}
        &
        \textbf{Temporal treatment}
        &
        \textbf{Role in comparison}
        \\

        \midrule

        Zero rainfall
        &
        None
        &
        None
        &
        Dry or low-rainfall reference
        \\

        Day-of-year climatology
        &
        Cell-specific mean field
        &
        Seasonal reference
        &
        Training-period climatology
        \\

        Persistence
        &
        Previous field retained
        &
        One-day continuation
        &
        Short-term continuity reference
        \\

        Three-day mean
        &
        Cell-specific rainfall fields
        &
        Recent three-day mean
        &
        Short-window smoothed reference
        \\

        Seven-day mean
        &
        Cell-specific rainfall fields
        &
        Full input-period mean
        &
        Full-window smoothed reference
        \\

        Ridge
        &
        Flattened field
        &
        All seven days used as predictors
        &
        Linear statistical model
        \\

        ExtraTrees
        &
        Flattened field
        &
        All seven days used as predictors
        &
        Non-linear tree ensemble
        \\

        CNN-only
        &
        Local convolutions
        &
        Days stacked as channels
        &
        Spatial model without recurrence
        \\

        FC-LSTM
        &
        Field flattened by day
        &
        Fully connected recurrence
        &
        Recurrent model without convolution
        \\

        TimeDistributed-CNN--ConvLSTM
        &
        Local convolutions
        &
        Convolutional recurrence
        &
        Combined spatial--temporal model
        \\

        \bottomrule

    \end{tabularx}
\end{table}


\subsection{Model training}
\label{subsec:model_training}

Ridge, ExtraTrees and all neural models were fitted to a logarithmically
transformed rainfall target, and predictions were returned to the original
rainfall scale before evaluation:

\begin{equation}
    \begin{aligned}
        \mathbf{Y}^{*}
        &=
        \log\left(1+\mathbf{Y}\right), \\
        \widehat{\mathbf{Y}}
        &=
        \max\left[
            0,
            \exp\left(\widehat{\mathbf{Y}}^{*}\right)-1
        \right].
    \end{aligned}
    \label{eq:target_transformation}
\end{equation}

All reported performance measures were therefore calculated in
mm day$^{-1}$ rather than in logarithmic space.

One prespecified configuration was evaluated for each model rather than
conducting exhaustive hyperparameter optimisation. The configurations were
selected through preliminary validation experiments and were then applied
consistently across the four cities and two input formulations. The comparison
therefore concerns the implemented configurations and should not be interpreted
as a definitive ranking of the broader model families.

The neural models were trained using the Adam optimiser with an initial
learning rate of $10^{-3}$, mean squared error loss, a batch size of 16 and a
maximum of 80 epochs. Early stopping monitored validation loss with a
patience of 12 epochs and restored the weights associated with the lowest
validation loss. The learning rate was reduced by a factor of 0.5 after six
epochs without validation improvement, subject to a minimum learning rate of
$10^{-5}$.

Each neural architecture was fitted separately for every city, input
formulation and random seed. The five seeds were 1234, 2025, 777, 42 and
9090.

Main neural-model performance was summarised using the mean and standard
deviation across the five runs. Model weights were updated using the training
samples. Validation data did not enter the gradient updates, but they were used
for early stopping, learning-rate reduction, restoration of the best-performing
epoch, preliminary configuration selection and selection of the fitted run used
for the post-hoc XAI analyses. The test period was withheld from preprocessing,
model fitting, early stopping, configuration selection and run selection.
Robust-scaler parameters, day-of-year climatological fields and city-specific
high-rainfall thresholds were estimated from training data only. The archived
runs recorded TensorFlow 2.20.0, NumPy 2.0.2 and pandas 2.2.2; the remaining
recorded training and software settings are listed in Supplementary Table~S4c.


\subsection{Performance evaluation}
\label{subsec:performance_evaluation}

Performance was evaluated after predictions had been inverse-transformed to
mm day$^{-1}$. Let $N$ denote the number of test days, and let $H$ and $W$
denote the spatial dimensions of the city-specific rainfall field.

Complete-field RMSE was calculated across all test days and grid cells:

\begin{equation}
    \mathrm{RMSE}_{\mathrm{field}}
    =
    \sqrt{
        \frac{1}{NHW}
        \sum_{n=1}^{N}
        \sum_{i=1}^{H}
        \sum_{j=1}^{W}
        \left(
            Y_{n,i,j}-\widehat{Y}_{n,i,j}
        \right)^{2}
    }.
    \label{eq:field_rmse}
\end{equation}

The observed and predicted domain-mean rainfall values for test day $n$ were

\begin{equation}
    \overline{Y}_{n}
    =
    \frac{1}{HW}
    \sum_{i=1}^{H}
    \sum_{j=1}^{W}Y_{n,i,j},
    \qquad
    \widehat{\overline{Y}}_{n}
    =
    \frac{1}{HW}
    \sum_{i=1}^{H}
    \sum_{j=1}^{W}\widehat{Y}_{n,i,j}.
    \label{eq:domain_means}
\end{equation}

Domain-mean rainfall is the arithmetic mean of precipitation across all
$H\times W$ retained grid cells on a given day. It is not a spatial rainfall
total or an estimate of rainfall volume over the domain. Domain-mean rainfall
RMSE was calculated as

\begin{equation}
    \mathrm{RMSE}_{\mathrm{domain}}
    =
    \sqrt{
        \frac{1}{N}
        \sum_{n=1}^{N}
        \left(
            \overline{Y}_{n}-\widehat{\overline{Y}}_{n}
        \right)^{2}
    }.
    \label{eq:domain_rmse}
\end{equation}

Domain-mean MAE was the mean absolute difference between predicted and observed
daily domain means. Mean bias was the average predicted minus observed
domain-mean rainfall; negative values therefore indicate underestimation.

Spatial anomalies were obtained by removing the daily domain mean from both the
observed and predicted fields:

\begin{equation}
    Y'_{n,i,j}
    =
    Y_{n,i,j}-\overline{Y}_{n},
    \qquad
    \widehat{Y}'_{n,i,j}
    =
    \widehat{Y}_{n,i,j}-\widehat{\overline{Y}}_{n}.
    \label{eq:spatial_anomalies}
\end{equation}

Spatial-anomaly RMSE was then calculated as

\begin{equation}
    \mathrm{RMSE}_{\mathrm{spatial}}
    =
    \sqrt{
        \frac{1}{NHW}
        \sum_{n=1}^{N}
        \sum_{i=1}^{H}
        \sum_{j=1}^{W}
        \left(
            Y'_{n,i,j}-\widehat{Y}'_{n,i,j}
        \right)^{2}
    }.
    \label{eq:spatial_rmse}
\end{equation}

This measure evaluates the within-field distribution of rainfall independently
of the predicted domain-mean amount.

RMSE skill and the lower-RMSE na\"ive reference used in the summary comparison
were defined jointly as

\begin{equation}
    \begin{aligned}
        \mathrm{SS}_{\mathrm{RMSE}}
        &=
        1-
        \frac{\mathrm{RMSE}_{\mathrm{model}}}
             {\mathrm{RMSE}_{\mathrm{reference}}}, \\
        \mathrm{RMSE}_{\mathrm{reference}}
        &=
        \min\left(
            \mathrm{RMSE}_{\mathrm{persistence}},
            \mathrm{RMSE}_{\mathrm{climatology}}
        \right).
    \end{aligned}
    \label{eq:forecast_skill}
\end{equation}

Skill was also calculated separately relative to persistence and grid
climatology. Positive values indicate lower RMSE than the selected reference,
negative values indicate higher RMSE and zero indicates equal RMSE.

A city-specific high-rainfall threshold was calculated as the 90th percentile
of training-period observed domain-mean rainfall:

\begin{equation}
    q_{0.90}^{(c)}
    =
    Q_{0.90}\left(
        \overline{Y}_{\mathrm{train}}^{(c)}
    \right),
    \label{eq:high_rainfall_threshold}
\end{equation}

where $c$ denotes the city. A test day was classified as a high-rainfall day
when its observed domain-mean rainfall equalled or exceeded the corresponding
training-period threshold. The classification therefore identifies days with
high rainfall averaged across the retained city domain and does not identify a
localised cell-level rainfall extreme.

Rainfall magnitude on these days was evaluated using domain-mean rainfall RMSE,
MAE and bias. Detection performance was evaluated using probability of
detection, false-alarm ratio, precision, critical success index, F1 score and
frequency bias. The city-specific threshold and number of observed
high-rainfall test days were reported alongside these scores. Because the
thresholds and test-period event frequencies differed among cities,
categorical scores were interpreted primarily among models within the same city.

Rainfall-field structure was described using the complete June--September
1998--2020 record. Total cell--day rainfall variance was separated into the
variance of the daily domain means and the mean within-field spatial variance.
Short-term spatial continuity was measured by removing the domain mean from
each field and calculating the Pearson correlation across grid cells for each
valid pair of consecutive days within the same year. Pairs were excluded when
either spatial-anomaly field had a spatial standard deviation of $10^{-8}$ or
less. The median valid lag-1 correlation was retained for each city.


\subsection{Post-hoc explainable artificial intelligence}
\label{subsec:xai_methods}

Post-hoc explainable artificial intelligence (XAI) analyses were conducted for
the \texttt{RAIN\_HIST\_ATM} formulation because it supplied the neural models
with both atmospheric conditions and recent rainfall fields. The analyses were
applied to FC-LSTM and TimeDistributed-CNN--ConvLSTM. For each city and
architecture, one fitted run was selected using the lowest validation-period
domain-mean rainfall RMSE. Test-period performance was not used for model
selection. The post-hoc XAI results therefore describe one
validation-selected fitted network per city and architecture rather than
averages across the five training seeds.

The analyses comprised grouped permutation sensitivity, temporal occlusion,
spatial-cell occlusion and Grad-CAM. Changes were evaluated for both
domain-mean rainfall RMSE and spatial-anomaly RMSE unless otherwise stated.

\subsubsection{Grouped permutation sensitivity}

The predictors were organised into six groups: past rainfall, wind, relative
humidity, cloud cover, surface pressure and evaporation. The wind group
contained zonal wind, meridional wind and wind speed. The static grid-position
channels were excluded from the meteorological interpretation.

For each group, the complete spatiotemporal block of all variables in that group
was permuted across test samples. The internal arrangement of lags, grid cells
and grouped channels within each permuted sample was retained. Each group was
permuted five times. Grouped permutation sensitivity was calculated as the
difference between RMSE after permutation and the unperturbed test RMSE:

\begin{equation}
    \Delta\mathrm{RMSE}_{g}
    =
    \mathrm{RMSE}_{g,\mathrm{permuted}}
    -
    \mathrm{RMSE}_{\mathrm{base}},
    \label{eq:permutation_sensitivity}
\end{equation}

where $g$ denotes the predictor group. Positive values indicate that
permutation increased prediction error, while negative values indicate that
error decreased after permutation. The mean and standard deviation across the
five repetitions describe variation in the permutation procedure and are not
confidence intervals across independently fitted models. Grouped scores
represent reliance on the information contained in a complete predictor group
and should not be interpreted as per-variable importance or causal effects,
particularly where predictors are correlated
\citep{Strobl2008,Gregorutti2015,Gregorutti2017}.

\subsubsection{Temporal occlusion}

Temporal sensitivity was examined by replacing each input day from $t-6$ to $t$
separately. The replacement field was the cell- and channel-specific median
calculated across all training sequences and all lag positions. The increase in
RMSE relative to the unmodified test inputs was recorded for each occluded day.

Median replacement avoids introducing arbitrary zero-valued atmospheric fields.
It does not, however, guarantee that the modified seven-day sequence lies
within the joint distribution of the training data.

\subsubsection{Spatial-cell occlusion}

Spatial-cell sensitivity was evaluated on observed high-rainfall test days. For
each retained grid cell, all channels at that location were replaced across the
complete seven-day input sequence. The replacement value for each channel was
the median calculated for that cell across all training samples and lag
positions. Because the two static position channels are constant at a given
cell, their medians equalled their original values and they remained unchanged
numerically. The resulting change in domain-mean rainfall RMSE and
spatial-anomaly RMSE was assigned to the occluded cell.

Positive values indicate that replacing information at the cell increased
prediction error. The analysis describes sensitivity at the retained IMDAA
grid-cell scale and does not provide sub-grid or ward-level spatial attribution.

\subsubsection{Grad-CAM}

Grad-CAM was applied only to the validation-selected ConvLSTM models. For each event, gradients were calculated
for the predicted domain-mean rainfall after inverse transformation to the
original rainfall unit:

\begin{equation}
    s
    =
    \frac{1}{HW}
    \sum_{i=1}^{H}
    \sum_{j=1}^{W}
    \max
    \left[
        0,
        \exp\left(\widehat{Y}^{*}_{i,j}\right)-1
    \right],
    \label{eq:gradcam_score}
\end{equation}

where $\widehat{Y}^{*}$ is the ConvLSTM output in logarithmic space. Gradients
of $s$ with respect to the feature maps of the final ConvLSTM layer were
spatially averaged to obtain channel weights. The weighted feature maps were
summed and passed through a rectified linear unit to retain positive
gradient-weighted activation.

Grad-CAM was calculated for up to the 30 wettest observed high-rainfall test
days in each city. Each event-level map was independently normalised to the
interval $[0,1]$, after which the event-level mean and standard deviation were
calculated. The resulting composites show relative positive activation within
each selected model. Their absolute magnitudes cannot be compared
quantitatively across cities.

All spatial XAI outputs were displayed on the retained $0.12^{\circ}$ IMDAA
grid cells without interpolation. Administrative boundaries were included only
for cartographic reference and were not used by the forecasting models. The
post-hoc XAI analyses describe fitted-model sensitivity and activation under
specific perturbations or gradients. They do not identify causal atmospheric
processes.


\section{Results}
\label{sec:results}

All performance measures were calculated for the independent 2018--2020 test
period after inverse transformation of the predicted rainfall fields to
mm\,day$^{-1}$. Results for CNN-only, FC-LSTM and ConvLSTM are reported as
means and between-seed standard deviations across five independently trained
runs. Zero-rainfall, climatology, persistence, moving-average, ridge-regression
and ExtraTrees forecasts were calculated or fitted once.


\subsection{Rainfall-field variability and short-term spatial continuity}
\label{subsec:rainfall_structure}

The four city domains differed in the relative contributions of domain-mean
variation and within-field spatial variation, as well as in the short-term
continuity of their spatial-anomaly fields. The domain-mean component accounted
for 45.1\% of total cell--day rainfall variance in Bengaluru, compared with
59.7\% in Delhi, 68.6\% in Kolkata and 73.8\% in Mumbai
(Table~\ref{tab:rainfall_structure}). The corresponding contribution of
within-field spatial variation declined from 54.9\% in Bengaluru to 26.2\% in
Mumbai.

The variance decomposition did not indicate whether the spatial configuration
of rainfall persisted between consecutive days. Median lag-1 correlations
between successive daily spatial-anomaly fields were 0.097 in Bengaluru, 0.069
in Delhi and 0.089 in Kolkata. Mumbai differed from the other city domains,
with a median lag-1 correlation of 0.392. Thus, within-field spatial variation
accounted for a smaller share of total rainfall variance in Mumbai, but the
configuration of those spatial anomalies was more persistent from one day to
the next.

The training-period 90th-percentile ($P_{90}$) thresholds for domain-mean
rainfall were 9.37 mm\,day$^{-1}$ in Bengaluru, 15.62 mm\,day$^{-1}$ in Delhi,
22.61 mm\,day$^{-1}$ in Kolkata and 29.19 mm\,day$^{-1}$ in Mumbai. These
thresholds identified 59, 43, 38 and 64 high-rainfall test days, corresponding
to test-period event rates of 17.1\%, 12.5\%, 11.0\% and 18.6\%, respectively.
The test-period exceedance rates therefore differed from the nominal 10\%
training-period threshold. Additional training- and test-period rainfall
distribution summaries are reported in Supplementary Table~S5a, while
standard-deviation-normalised domain-mean rainfall RMSE values are reported in
Supplementary Table~S5b.

\begin{table}[H]
    \centering
    \footnotesize
    \setlength{\tabcolsep}{2.3pt}
    \renewcommand{\arraystretch}{1.12}

    \caption{Rainfall-field variance, short-term spatial continuity and
    high-rainfall thresholds.}
    \label{tab:rainfall_structure}

    \begin{tabular}{lccccc}
        \toprule

        \textbf{City}
        &
        \textbf{\begin{tabular}[c]{@{}c@{}}Domain-mean\\variance (\%)\end{tabular}}
        &
        \textbf{\begin{tabular}[c]{@{}c@{}}Spatial\\variance (\%)\end{tabular}}
        &
        \textbf{\begin{tabular}[c]{@{}c@{}}Median lag-1\\spatial correlation\end{tabular}}
        &
        \textbf{\begin{tabular}[c]{@{}c@{}}Training $P_{90}$\\(mm\,day$^{-1}$)\end{tabular}}
        &
        \textbf{\begin{tabular}[c]{@{}c@{}}High-rainfall test\\days, $n$ (\%)\end{tabular}}
        \\

        \midrule

        Bengaluru & 45.1 & 54.9 & 0.097 & 9.37  & 59 (17.1) \\
        Delhi      & 59.7 & 40.3 & 0.069 & 15.62 & 43 (12.5) \\
        Kolkata    & 68.6 & 31.4 & 0.089 & 22.61 & 38 (11.0) \\
        Mumbai     & 73.8 & 26.2 & 0.392 & 29.19 & 64 (18.6) \\

        \bottomrule
    \end{tabular}

    \begin{minipage}{0.95\textwidth}
        \vspace{3pt}
        \footnotesize
        \textit{Note:}
        Domain-mean and spatial variance shares were calculated from all
        June--September daily fields from 1998 to 2020 using population
        variances and sum to 100\% within each city. Lag-1 spatial continuity
        is the median Pearson correlation across cells for valid
        consecutive-day spatial-anomaly pairs within the same year; pairs with
        near-zero spatial variance were excluded. High-rainfall thresholds
        were estimated from training-period domain-mean rainfall only.
    \end{minipage}
\end{table}


\subsection{Benchmark performance and RMSE skill}
\label{subsec:benchmark_results}

\begin{figure}[p]
    \centering
    \includegraphics[
        width=\textwidth,
        height=0.82\textheight,
        keepaspectratio
    ]{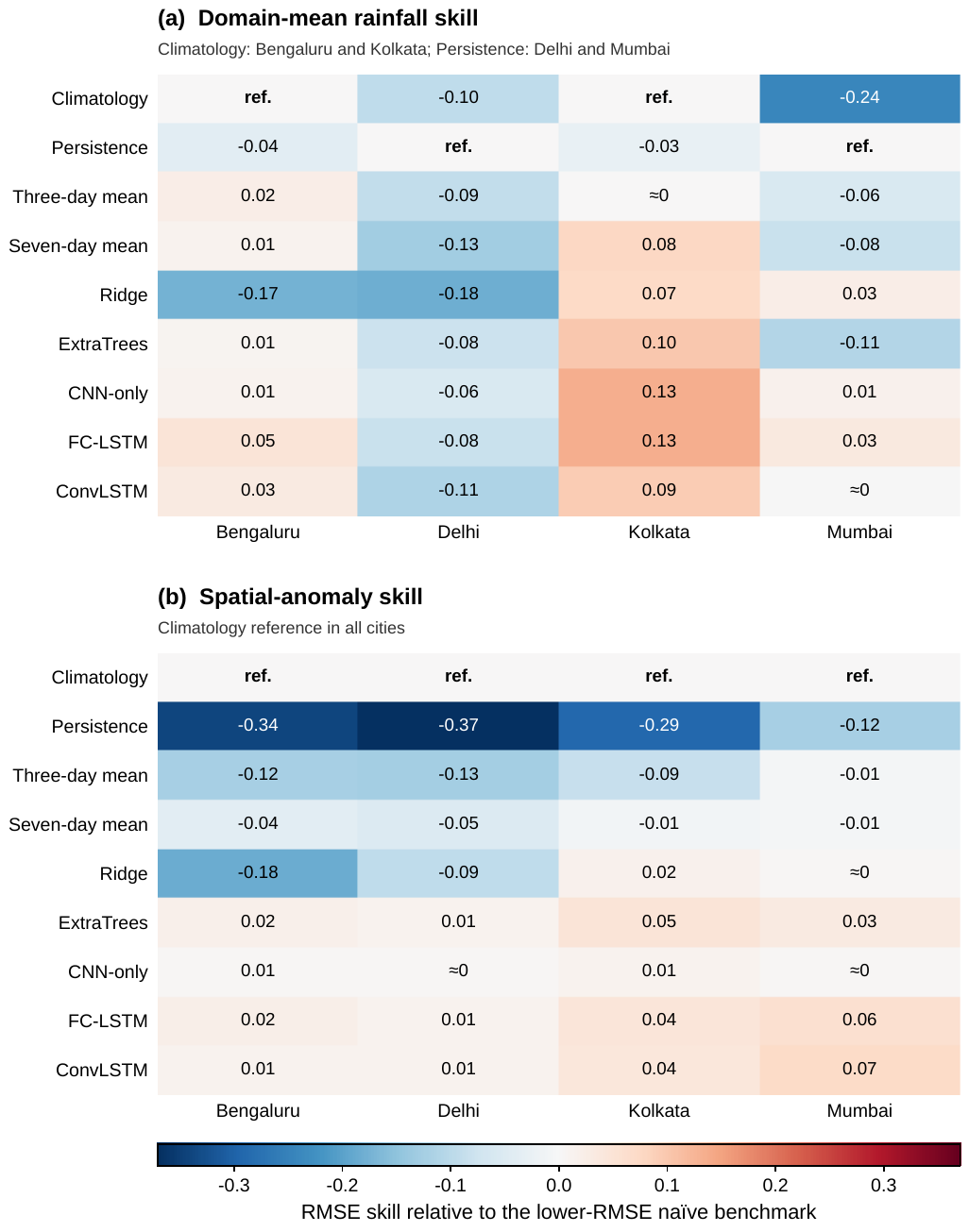}

    \caption{Test-period RMSE skill under the rainfall-history-plus-atmospheric
    (\texttt{RAIN\_HIST\_ATM}) input formulation relative to the lower-RMSE
    na\"ive benchmark. Panel~(a) reports skill for domain-mean rainfall, using
    grid climatology as the reference in Bengaluru and Kolkata and persistence
    in Delhi and Mumbai. Panel~(b) reports spatial-anomaly skill using grid
    climatology as the reference in all four cities. Skill was calculated as
    \(1-\mathrm{RMSE}_{\mathrm{model}}/
    \mathrm{RMSE}_{\mathrm{reference}}\). Positive values indicate lower RMSE
    than the selected benchmark, negative values indicate higher RMSE, and
    values with absolute skill below \(0.005\) are displayed as \(\approx 0\).
    Neural-model values are based on mean RMSE across five independently
    trained seeds, whereas deterministic reference and non-neural models were
    fitted once. The zero-rainfall benchmark is omitted from the heatmap and is
    reported in Supplementary Tables~S1a and~S1b.}

    \label{fig:benchmark_skill}
\end{figure}

No forecasting method produced the numerically lowest error across every city
and evaluation metric. Under \texttt{RAIN\_HIST\_ATM}, FC-LSTM produced the
lowest mean domain-mean rainfall RMSE in Bengaluru, Kolkata and Mumbai, whereas
persistence produced the lowest value in Delhi
(Figure~\ref{fig:benchmark_skill};
Tables~\ref{tab:benchmark_amount_summary}
and~\ref{tab:benchmark_spatial_summary}). For the summary skill comparison,
grid climatology was the lower-RMSE na\"ive reference in Bengaluru and Kolkata,
while persistence was the lower-RMSE reference in Delhi and Mumbai.

Relative to the selected references, FC-LSTM produced domain-mean rainfall
skill scores of 0.048 in Bengaluru, 0.134 in Kolkata and 0.035 in Mumbai. In
Kolkata, the FC-LSTM and CNN-only RMSE values were nearly indistinguishable
numerically at $9.28\pm0.17$ and $9.29\pm0.11$ mm\,day$^{-1}$, respectively.
No fitted statistical, tree-based or neural model improved on persistence for
domain-mean rainfall in Delhi.

Spatial-anomaly RMSE values were more closely grouped. Grid climatology was the
lower-RMSE na\"ive reference in all four cities. The numerically lowest
spatial-anomaly RMSE corresponded to selected-reference skill scores of 0.020
for FC-LSTM in Bengaluru, 0.014 for ExtraTrees in Delhi, 0.047 for ExtraTrees in
Kolkata and 0.071 for ConvLSTM in Mumbai. FC-LSTM and ExtraTrees differed by
only 0.01 mm\,day$^{-1}$ in Bengaluru. ExtraTrees, FC-LSTM and ConvLSTM were
separated by no more than 0.05 mm\,day$^{-1}$ in Delhi and Kolkata. In Mumbai,
ConvLSTM produced the numerically lowest spatial-anomaly RMSE
($10.45\pm0.15$ mm\,day$^{-1}$), but its difference from FC-LSTM
($10.58\pm0.29$ mm\,day$^{-1}$) was small and was not subjected to paired
uncertainty testing.

For complete-field RMSE, FC-LSTM produced the numerically lowest mean in
Bengaluru ($9.30\pm0.08$ mm\,day$^{-1}$), Kolkata
($11.87\pm0.14$ mm\,day$^{-1}$) and Mumbai
($17.77\pm0.63$ mm\,day$^{-1}$), while CNN-only produced the lowest mean in
Delhi ($21.10\pm0.31$ mm\,day$^{-1}$). Complete-field RMSE, domain-mean
rainfall MAE and bias for all models and both input formulations are reported in
Supplementary Table~S1a. Skill scores relative to persistence and climatology
separately are reported in Supplementary Table~S1b.

\begin{table}[H]
    \centering
    \footnotesize
    \setlength{\tabcolsep}{3.5pt}
    \caption{Numerically lowest domain-mean rainfall RMSE and skill relative
    to the selected lower-RMSE na\"ive reference under
    \texttt{RAIN\_HIST\_ATM}.}
    \label{tab:benchmark_amount_summary}
    \begin{tabularx}{\textwidth}{@{}l>{\raggedright\arraybackslash}Xc>{\raggedright\arraybackslash}Xc@{}}
        \toprule
        \textbf{City} & \textbf{Model} & \textbf{RMSE} &
        \textbf{Selected na\"ive reference} &
        \textbf{$SS_{\mathrm{RMSE}}$} \\
        \midrule
        Bengaluru & FC-LSTM & $5.98\pm0.13$ & Grid climatology & 0.048 \\
        Delhi & Persistence & 14.54 & Persistence & 0.000 \\
        Kolkata & FC-LSTM & $9.28\pm0.17$ & Grid climatology & 0.134 \\
        Mumbai & FC-LSTM & $14.27\pm0.57$ & Persistence & 0.035 \\
        \bottomrule
    \end{tabularx}
    \begin{minipage}{0.94\textwidth}
        \vspace{3pt}\footnotesize
        \textit{Note:} RMSE is in mm\,day$^{-1}$. The selected reference is the
        lower-RMSE forecast between persistence and grid climatology for the
        stated city and metric. Rankings are numerical; close differences were
        not subjected to paired uncertainty testing.
    \end{minipage}
\end{table}

\begin{table}[H]
    \centering
    \footnotesize
    \setlength{\tabcolsep}{3.5pt}
    \caption{Numerically lowest spatial-anomaly RMSE and skill relative to the
    selected lower-RMSE na\"ive reference under
    \texttt{RAIN\_HIST\_ATM}.}
    \label{tab:benchmark_spatial_summary}
    \begin{tabularx}{\textwidth}{@{}l>{\raggedright\arraybackslash}Xc>{\raggedright\arraybackslash}Xc@{}}
        \toprule
        \textbf{City} & \textbf{Model} & \textbf{RMSE} &
        \textbf{Selected na\"ive reference} &
        \textbf{$SS_{\mathrm{RMSE}}$} \\
        \midrule
        Bengaluru & FC-LSTM & $7.13\pm0.01$ & Grid climatology & 0.020 \\
        Delhi & ExtraTrees & 14.29 & Grid climatology & 0.014 \\
        Kolkata & ExtraTrees & 7.38 & Grid climatology & 0.047 \\
        Mumbai & ConvLSTM & $10.45\pm0.15$ & Grid climatology & 0.071 \\
        \bottomrule
    \end{tabularx}
    \begin{minipage}{0.94\textwidth}
        \vspace{3pt}\footnotesize
        \textit{Note:} RMSE is in mm\,day$^{-1}$. Grid climatology was the
        lower-RMSE na\"ive spatial reference in all four cities. Rankings are
        numerical; close differences were not subjected to paired uncertainty
        testing.
    \end{minipage}
\end{table}


\subsection{Contribution of recent rainfall fields}
\label{subsec:rainfall_history_results}

Adding previous rainfall fields did not improve the neural models uniformly
(Table~\ref{tab:rain_history_change}). Changes were negligible in Bengaluru.
In Delhi, domain-mean rainfall RMSE declined by approximately
0.30 mm\,day$^{-1}$ for CNN-only and FC-LSTM but changed little for ConvLSTM.
The changes in spatial-anomaly RMSE were no greater than
0.03 mm\,day$^{-1}$ for any of the three architectures.

In Kolkata, the principal improvement occurred for CNN-only, whose domain-mean
rainfall RMSE and spatial-anomaly RMSE declined by 0.79 and
1.05 mm\,day$^{-1}$, respectively. Changes for FC-LSTM and ConvLSTM were small:
domain-mean rainfall RMSE declined by 0.09 mm\,day$^{-1}$ for FC-LSTM and
increased by 0.05 mm\,day$^{-1}$ for ConvLSTM, while spatial-anomaly RMSE
declined by 0.02 mm\,day$^{-1}$ for both models.

Mumbai was the only city in which rainfall history lowered both domain-mean
rainfall RMSE and spatial-anomaly RMSE for all three neural architectures. The
reductions ranged from 1.09 to 1.46 mm\,day$^{-1}$ for domain-mean rainfall and
from 0.46 to 0.90 mm\,day$^{-1}$ for spatial anomalies. Mumbai also had the
highest median lag-1 spatial-anomaly correlation, whereas the other city
domains had correlations below 0.10. This cross-city pattern is consistent with
a greater contribution from recent rainfall where day-to-day spatial
continuity was stronger, but it does not establish spatial continuity as the
cause of the model differences.

ConvLSTM did not produce the lowest domain-mean rainfall RMSE under either input
formulation. Its only numerically lowest principal result was the Mumbai
spatial-anomaly RMSE under \texttt{RAIN\_HIST\_ATM}.

\begin{table}[H]
    \centering
    \small
    \setlength{\tabcolsep}{6pt}
    \renewcommand{\arraystretch}{1.15}

    \caption{Change in neural-model RMSE after adding previous rainfall
    fields.}
    \label{tab:rain_history_change}

    \begin{tabular}{llcc}
        \toprule

        \textbf{City}
        &
        \textbf{Model}
        &
        \textbf{\begin{tabular}[c]{@{}c@{}}Change in domain-mean\\rainfall RMSE (mm\,day$^{-1}$)\end{tabular}}
        &
        \textbf{\begin{tabular}[c]{@{}c@{}}Change in spatial-anomaly\\RMSE (mm\,day$^{-1}$)\end{tabular}}
        \\

        \midrule

        Bengaluru & CNN-only & $\approx 0$ & $+0.02$ \\
                  & FC-LSTM  & $-0.01$ & $\approx 0$ \\
                  & ConvLSTM & $+0.01$ & $+0.02$ \\

        \addlinespace

        Delhi     & CNN-only & $-0.31$ & $+0.03$ \\
                  & FC-LSTM  & $-0.30$ & $\approx 0$ \\
                  & ConvLSTM & $+0.01$ & $-0.01$ \\

        \addlinespace

        Kolkata   & CNN-only & $-0.79$ & $-1.05$ \\
                  & FC-LSTM  & $-0.09$ & $-0.02$ \\
                  & ConvLSTM & $+0.05$ & $-0.02$ \\

        \addlinespace

        Mumbai    & CNN-only & $-1.43$ & $-0.90$ \\
                  & FC-LSTM  & $-1.46$ & $-0.46$ \\
                  & ConvLSTM & $-1.09$ & $-0.66$ \\

        \bottomrule
    \end{tabular}

    \begin{minipage}{0.94\textwidth}
        \vspace{3pt}
        \footnotesize
        \textit{Note:}
        Change is calculated as
        $\mathrm{RMSE}_{\texttt{RAIN\_HIST\_ATM}}-
        \mathrm{RMSE}_{\texttt{ATM\_ONLY}}$.
        Negative values indicate lower error after rainfall history was added.
        Values are differences between five-seed mean RMSE values; absolute
        changes below 0.005 mm\,day$^{-1}$ are displayed as $\approx 0$.
    \end{minipage}
\end{table}


\subsection{High-rainfall magnitude and detection}
\label{subsec:high_rain_results}

Performance on high-rainfall days differed markedly from overall test-period
performance. FC-LSTM and ConvLSTM had negative high-rainfall biases in all four
cities, indicating systematic underestimation on days at or above the
city-specific training-period $P_{90}$ threshold
(Table~\ref{tab:high_rain_results}). The limitation was most pronounced in
Delhi, where FC-LSTM and ConvLSTM high-rainfall RMSE exceeded
43 mm\,day$^{-1}$ and mean biases were approximately $-29$ to
$-30$ mm\,day$^{-1}$, compared with a persistence RMSE of
33.77 mm\,day$^{-1}$ and bias of $-10.15$ mm\,day$^{-1}$. Persistence also
produced lower high-rainfall RMSE than either recurrent model in Bengaluru,
while the seven-day mean produced the lowest high-rainfall RMSE in Kolkata. In
Mumbai, FC-LSTM and persistence had nearly identical high-rainfall RMSE values,
but FC-LSTM had a substantially larger negative bias ($-21.53$ versus
$-7.06$ mm\,day$^{-1}$).

Threshold-based detection showed the same broad limitation. Persistence had the
highest probability of detection (POD) and critical success index (CSI) in
every city. Its POD ranged from 0.368 in Kolkata to 0.641 in Mumbai. FC-LSTM
mean POD ranged from 0.009 in Delhi to 0.288 in Mumbai, while ConvLSTM ranged
from zero in Delhi to 0.206 in Mumbai. None of the five Delhi ConvLSTM runs
predicted a domain-mean rainfall value above the high-rainfall threshold.

Several neural models had low false-alarm ratios (FAR), but these values did not
indicate effective event detection. Their frequency-bias values were
substantially below one, showing that they forecast far fewer high-rainfall days
than were observed. For neural models, FAR was defined only for seeds that
produced at least one predicted exceedance and was averaged across those seeds;
it was undefined for the remaining runs. A low FAR could therefore coexist with
severe underforecasting and must be interpreted together with POD, CSI and
frequency bias.

\clearpage
\begin{landscape}

\begin{table}[H]
    \centering
    \scriptsize
    \setlength{\tabcolsep}{3.7pt}
    \renewcommand{\arraystretch}{1.18}

    \caption{High-rainfall magnitude and detection under
    \texttt{RAIN\_HIST\_ATM}.}
    \label{tab:high_rain_results}

    \begin{tabular}{llccccccc}
        \toprule

        \textbf{City}
        &
        \textbf{Model}
        &
        \textbf{\begin{tabular}[c]{@{}c@{}}High-rainfall test\\days, $n$ (\%)\end{tabular}}
        &
        \textbf{\begin{tabular}[c]{@{}c@{}}High-rainfall\\RMSE\end{tabular}}
        &
        \textbf{Bias}
        &
        \textbf{POD}
        &
        \textbf{FAR$^{a}$}
        &
        \textbf{CSI}
        &
        \textbf{\begin{tabular}[c]{@{}c@{}}Frequency\\bias\end{tabular}}
        \\

        \midrule

        Bengaluru
        & Persistence & 59 (17.1) & 11.52 & $-5.61$
        & 0.407 & 0.593 & 0.255 & 1.000 \\

        & FC-LSTM & 59 (17.1) & $13.37\pm0.37$ & $-10.30$
        & 0.041 & 0.487 & 0.039 & 0.071 \\

        & ConvLSTM & 59 (17.1) & $13.56\pm0.48$ & $-10.59$
        & 0.024 & 0.167 & 0.024 & 0.031 \\

        \addlinespace

        Delhi
        & Persistence & 43 (12.5) & 33.77 & $-10.15$
        & 0.488 & 0.512 & 0.323 & 1.000 \\

        & FC-LSTM & 43 (12.5) & $43.38\pm0.54$ & $-28.67$
        & 0.009 & 0.000 & 0.009 & 0.009 \\

        & ConvLSTM & 43 (12.5) & $44.60\pm0.48$ & $-30.23$
        & 0.000 & -- & 0.000 & 0.000 \\

        \addlinespace

        Kolkata
        & Persistence & 38 (11.0) & 21.80 & $-11.25$
        & 0.368 & 0.622 & 0.230 & 0.974 \\

        & FC-LSTM & 38 (11.0) & $22.63\pm1.02$ & $-18.69$
        & 0.058 & 0.628 & 0.052 & 0.158 \\

        & ConvLSTM & 38 (11.0) & $23.48\pm1.21$ & $-19.41$
        & 0.084 & 0.408 & 0.075 & 0.184 \\

        \addlinespace

        Mumbai
        & Persistence & 64 (18.6) & 28.99 & $-7.06$
        & 0.641 & 0.359 & 0.471 & 1.000 \\

        & FC-LSTM & 64 (18.6) & $28.96\pm1.34$ & $-21.53$
        & 0.288 & 0.458 & 0.231 & 0.525 \\

        & ConvLSTM & 64 (18.6) & $30.95\pm2.86$ & $-23.56$
        & 0.206 & 0.369 & 0.172 & 0.328 \\

        \bottomrule
    \end{tabular}

    \begin{minipage}{0.94\linewidth}
        \vspace{3pt}
        \footnotesize
        \textit{Note:}
        RMSE and bias are in mm\,day$^{-1}$. High-rainfall days are observed
        test targets at or above the city-specific training-period $P_{90}$
        threshold for domain-mean rainfall. Neural RMSE values are means
        $\pm$ between-seed standard deviations; neural bias, POD, CSI and
        frequency bias are means across all five seeds. $^{a}$For neural
        models, FAR was summarised only across seeds that produced at least one
        predicted exceedance; a dash indicates that FAR was undefined for all
        five seeds. POD = probability of detection; FAR = false-alarm ratio;
        CSI = critical success index.
    \end{minipage}
\end{table}

Complete high-rainfall magnitude errors for all forecasting methods are
reported in Supplementary Table~S2a. The corresponding contingency counts and
categorical detection metrics are reported in Supplementary Table~S2b.

\end{landscape}
\clearpage


\subsection{Post-hoc XAI results}
\label{subsec:xai_results}

Post-hoc XAI was conducted for one validation-selected FC-LSTM and one
validation-selected ConvLSTM in each city under
\texttt{RAIN\_HIST\_ATM}. The selected run had the lowest validation-period
domain-mean rainfall RMSE for its city and architecture; test performance was
not used for selection. These analyses describe the behaviour of the selected
runs rather than the mean behaviour across all five seeds. The principal
grouped-permutation, temporal-occlusion and spatial-cell-occlusion results are
summarised in Table~\ref{tab:xai_summary}.


\subsubsection{Predictor-group sensitivity}
\label{subsubsec:permutation_results}

\begin{figure}[p]
    \centering
    \includegraphics[
        width=\textwidth,
        height=0.80\textheight,
        keepaspectratio
    ]{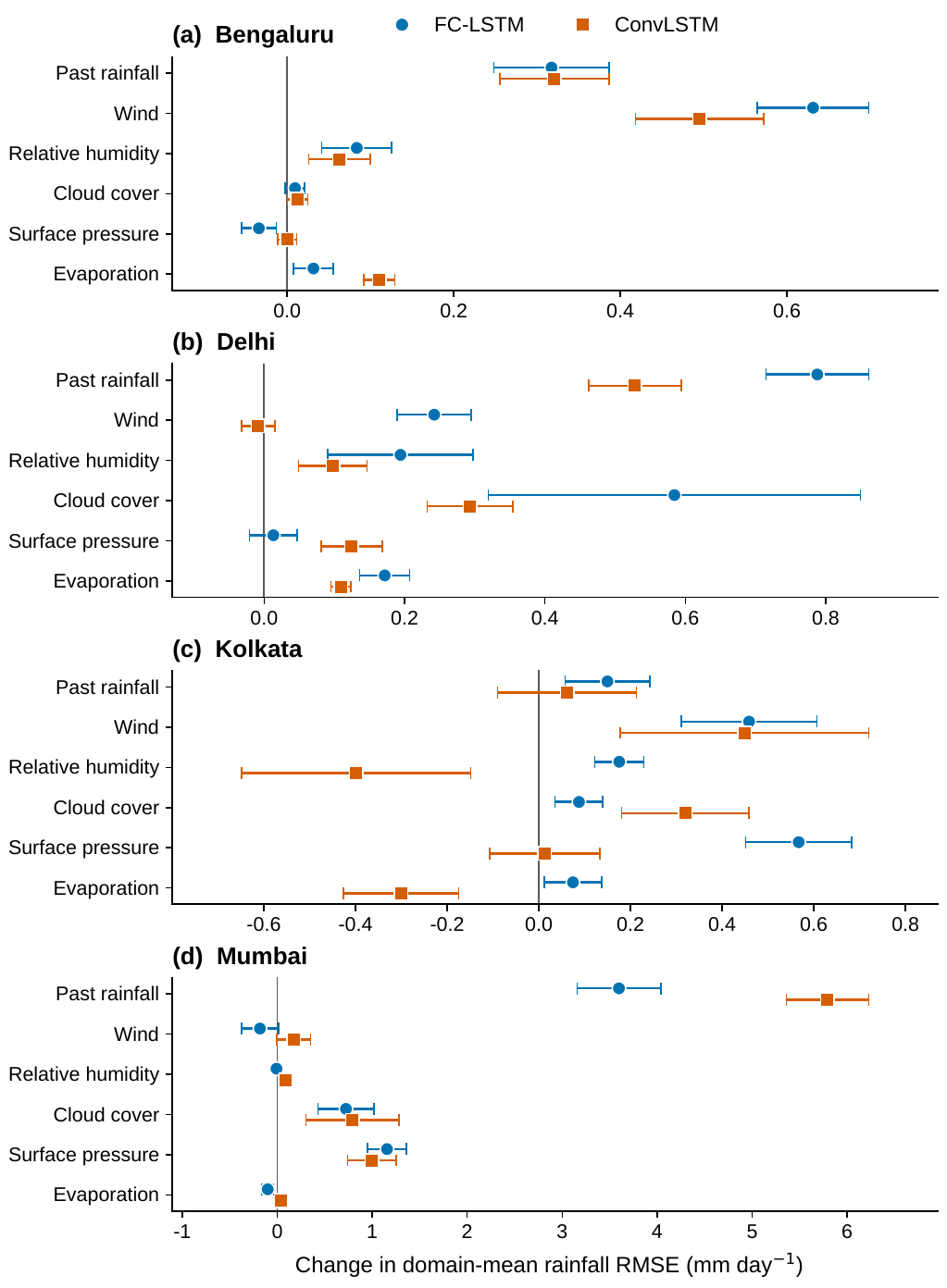}

    \caption{Grouped-permutation sensitivity for domain-mean rainfall
    prediction in the validation-selected FC-LSTM and ConvLSTM models. Each
    predictor group's complete seven-day spatial field was permuted across test
    samples while its internal temporal and spatial organisation was retained.
    Points show the mean change in domain-mean rainfall RMSE, and whiskers show
    the standard deviation across five permutation repetitions. Positive values
    indicate increased error after permutation; negative values indicate that
    permutation reduced test error and should not be interpreted as negative
    physical importance. City-specific horizontal scales are used to show the
    within-city patterns clearly. Absolute RMSE changes should not be compared
    as normalised measures of dependence across cities.}

    \label{fig:grouped_permutation_amount}
\end{figure}

Grouped permutation produced distinct city- and architecture-specific
sensitivity patterns (Figure~\ref{fig:grouped_permutation_amount}). Within the
selected Bengaluru models, wind was the predictor group whose permutation
produced the largest increase in domain-mean rainfall RMSE. The increase was
$0.63\pm0.07$ mm\,day$^{-1}$ for FC-LSTM and
$0.50\pm0.08$ mm\,day$^{-1}$ for ConvLSTM. Past rainfall ranked second for
both models, increasing RMSE by approximately 0.32 mm\,day$^{-1}$.

Within the selected Delhi models, past rainfall produced the largest increase
in domain-mean rainfall RMSE: $0.79\pm0.07$ mm\,day$^{-1}$ for FC-LSTM and
$0.53\pm0.07$ mm\,day$^{-1}$ for ConvLSTM. Cloud cover ranked second, with
increases of $0.58\pm0.27$ and $0.29\pm0.06$ mm\,day$^{-1}$, respectively.
Sensitivity to these predictor groups did not translate into lower domain-mean
rainfall RMSE than persistence.

Kolkata showed the clearest architecture-specific difference in predictor-group
ranking. The selected FC-LSTM was most sensitive to surface pressure, whose
permutation increased domain-mean rainfall RMSE by
$0.57\pm0.12$ mm\,day$^{-1}$, followed by wind at
$0.46\pm0.15$ mm\,day$^{-1}$. The selected ConvLSTM was most sensitive to wind
($0.45\pm0.27$ mm\,day$^{-1}$) and cloud cover
($0.32\pm0.14$ mm\,day$^{-1}$). Permuting past rainfall produced only a small
mean change for either architecture.

Within the selected Mumbai models, past rainfall was the predictor group whose
permutation produced the largest RMSE increase. Domain-mean rainfall RMSE
increased by $3.60\pm0.44$ mm\,day$^{-1}$ for FC-LSTM and
$5.79\pm0.43$ mm\,day$^{-1}$ for ConvLSTM, while spatial-anomaly RMSE increased
by 0.82 and 1.43 mm\,day$^{-1}$, respectively. Surface pressure ranked second,
increasing domain-mean rainfall RMSE by 1.16 mm\,day$^{-1}$ for FC-LSTM and
1.00 mm\,day$^{-1}$ for ConvLSTM.

Some predictor groups produced near-zero or negative changes. In those cases,
permutation did not degrade the selected model's test performance. Such values
may arise from predictor redundancy, correlations among inputs or variation
across the five permutations and are not interpreted as negative causal
effects.

The corresponding grouped-permutation effects on spatial-anomaly RMSE are
presented in Supplementary Figure~S1.


\subsubsection{Temporal sensitivity}
\label{subsubsec:temporal_results}

\begin{figure}[!t]
    \centering
    \includegraphics[
        width=\textwidth,
        height=0.72\textheight,
        keepaspectratio
    ]{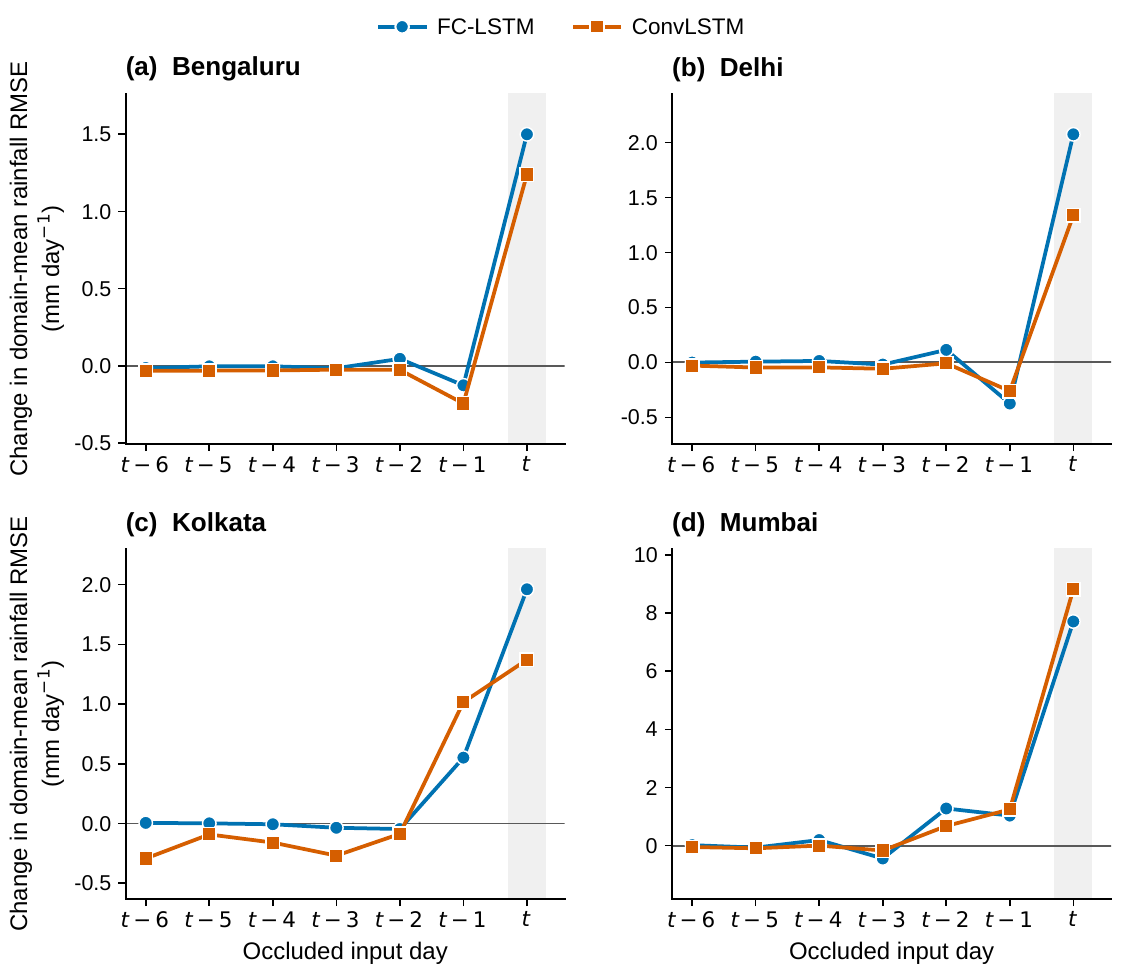}

    \caption{Temporal-occlusion sensitivity for domain-mean rainfall
    prediction in the validation-selected FC-LSTM and ConvLSTM models. Each
    input day from $t-6$ to $t$ was replaced individually with a cell- and
    channel-specific training-reference median field calculated across all
    training sequences and lag positions. Positive values indicate increased
    test RMSE after occlusion. The shaded band identifies the most recent input
    day, $t$. City-specific vertical scales are used to show each city's
    within-model lag pattern clearly. Absolute RMSE changes should not be
    compared as normalised measures of dependence across cities.}

    \label{fig:temporal_occlusion_amount}
\end{figure}

Replacing the most recent input day produced the largest increase in
domain-mean rainfall RMSE for every validation-selected model
(Figure~\ref{fig:temporal_occlusion_amount}). In Bengaluru, occluding day $t$
increased RMSE by 1.50 mm\,day$^{-1}$ for FC-LSTM and
1.24 mm\,day$^{-1}$ for ConvLSTM. The effects of days $t-6$ through $t-2$ were
close to zero, while replacement of $t-1$ marginally reduced RMSE.

Delhi showed a similar concentration on the most recent input day. Occluding
$t$ increased RMSE by 2.07 mm\,day$^{-1}$ for FC-LSTM and
1.34 mm\,day$^{-1}$ for ConvLSTM. Earlier input days produced small or negative
changes.

Kolkata showed additional sensitivity to $t-1$. For FC-LSTM, replacing $t$ and
$t-1$ increased RMSE by 1.96 and 0.55 mm\,day$^{-1}$, respectively. For
ConvLSTM, the corresponding increases were 1.37 and 1.02 mm\,day$^{-1}$.
Occluding earlier input days generally produced little deterioration.

The selected Mumbai models were sensitive to a broader set of recent lags.
Occluding $t$ increased RMSE by 7.71 mm\,day$^{-1}$ for FC-LSTM and
8.82 mm\,day$^{-1}$ for ConvLSTM. Days $t-1$ and $t-2$ also produced positive
increases: 1.03 and 1.28 mm\,day$^{-1}$ for FC-LSTM, and 1.25 and
0.68 mm\,day$^{-1}$ for ConvLSTM. The corresponding spatial-anomaly results
also showed positive deterioration after occlusion of the three most recent
input days.

The temporal-occlusion results show that the selected fitted models depended
primarily on recent input information. They do not represent different forecast
horizons: every model predicted the rainfall field one day ahead.

The corresponding temporal-occlusion effects on spatial-anomaly RMSE are
presented in Supplementary Figure~S2. Absolute RMSE changes are reported in the
figures but are interpreted within each city because rainfall variability and
baseline error differed across the four domains.


\subsubsection{Spatial-cell sensitivity and Grad-CAM activation}
\label{subsubsec:spatial_xai_results}

Spatial-cell occlusion produced heterogeneous within-city sensitivity patterns.
The maximum domain-mean rainfall RMSE increase after occluding one grid cell was
0.20 and 0.07 mm\,day$^{-1}$ for the selected Bengaluru FC-LSTM and ConvLSTM,
0.39 and 0.14 mm\,day$^{-1}$ in Delhi, 1.18 and
0.66 mm\,day$^{-1}$ in Kolkata, and 2.46 and
2.86 mm\,day$^{-1}$ in Mumbai. These absolute changes are reported
descriptively within each city and are not interpreted as directly comparable
sensitivity magnitudes across cities.

\begin{figure}[!t]
    \centering
    \includegraphics[
        width=0.98\textwidth,
        keepaspectratio
    ]{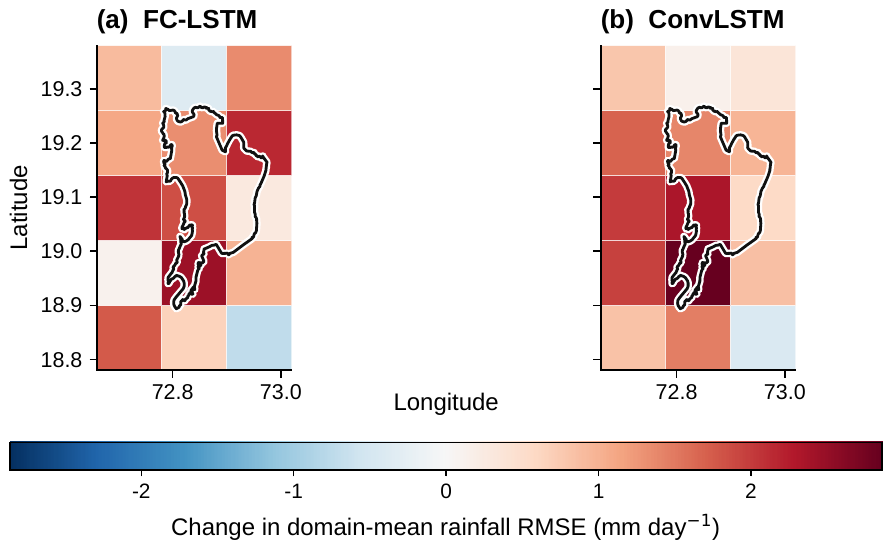}

    \caption{High-rainfall spatial-cell occlusion sensitivity for the
    validation-selected FC-LSTM and ConvLSTM models in Mumbai. Each coloured
    rectangle represents one retained $0.12^{\circ}$ modelling-grid cell.
    Values show the change in domain-mean rainfall RMSE when the dynamic
    predictor channels at that location across the seven input days were
    replaced with their cell- and channel-specific training medians. The static
    position channels retained their original values because their cell-specific
    medians were identical. The two panels use a common colour scale. Values
    are displayed at the retained $0.12^{\circ}$ grid-cell spacing without
    spatial interpolation. The dissolved outer boundary is included for
    geographic reference only and was not supplied to the models.}

    \label{fig:mumbai_cell_occlusion}
\end{figure}

For both selected Mumbai models, the maximum domain-mean rainfall RMSE increase
occurred at the grid cell centred at approximately $18.96^{\circ}$N,
$72.84^{\circ}$E (Figure~\ref{fig:mumbai_cell_occlusion}). Occluding this cell
increased ConvLSTM spatial-anomaly RMSE by 1.08 mm\,day$^{-1}$. This result
identifies the most sensitive retained cell within the selected Mumbai models;
it does not imply sub-grid spatial importance.

The corresponding domain-mean rainfall cell-occlusion results for Bengaluru,
Delhi and Kolkata are presented in Supplementary Figure~S3. Spatial-anomaly
cell-occlusion results for all four cities are presented in Supplementary
Figure~S4.

\begin{figure}[!t]
    \centering
    \includegraphics[
        width=0.98\textwidth,
        height=0.74\textheight,
        keepaspectratio
    ]{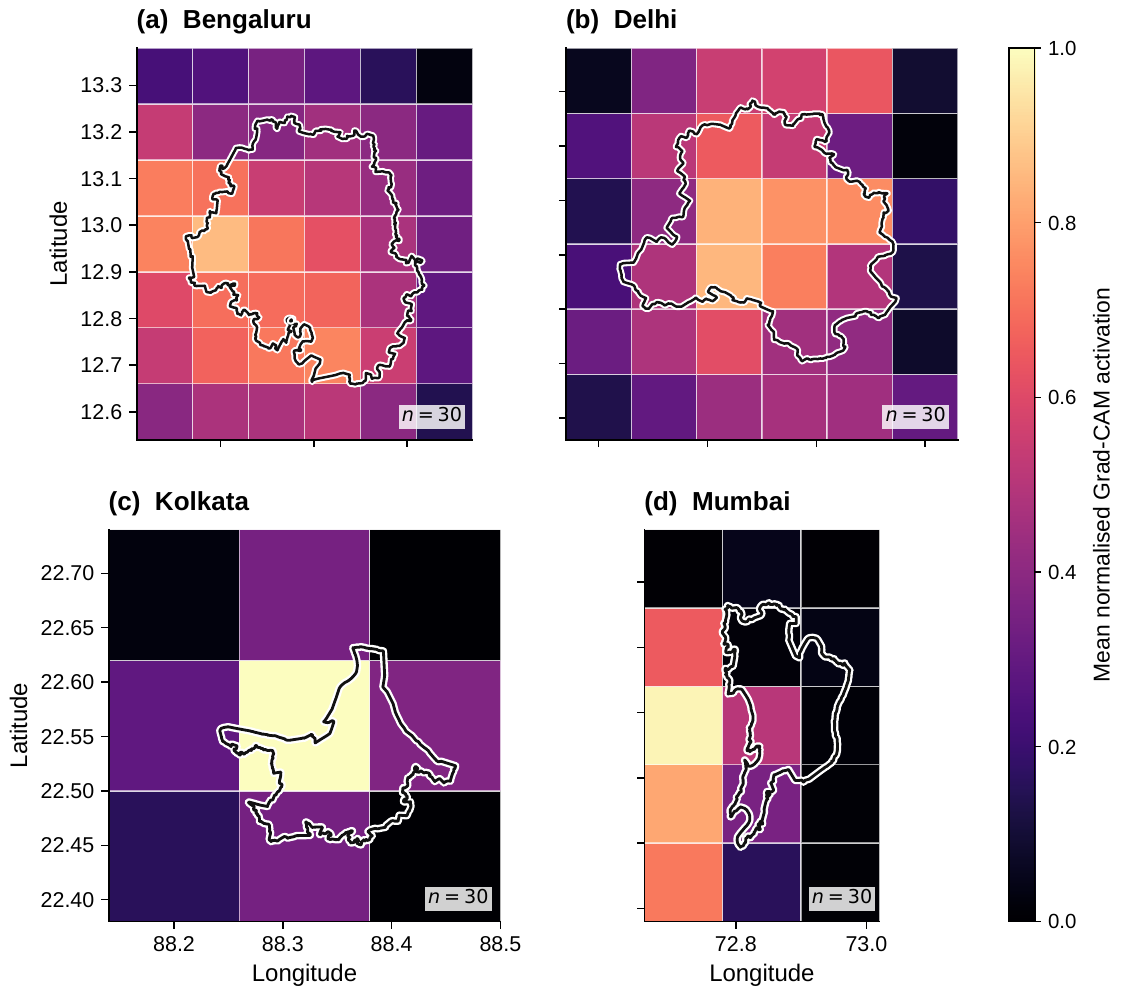}

    \caption{Mean normalised Grad-CAM activation for the validation-selected
    ConvLSTM models across the 30 wettest observed high-rainfall test events in
    each city. Each panel displays relative positive activation on the retained
    city-specific modelling grid without spatial interpolation. Event-level
    maps were independently normalised before averaging. The values therefore
    represent relative localisation within each fitted model and should not be
    interpreted as absolute contributions, causal atmospheric effects or
    directly comparable attribution magnitudes across cities. Dissolved outer
    boundaries are included for geographic reference only and were not supplied
    to the models. Grad-CAM was not calculated for FC-LSTM because the
    procedure used the final ConvLSTM feature maps.}

    \label{fig:gradcam_four_city}
\end{figure}

Grad-CAM composites were calculated for the 30 wettest observed high-rainfall
test events in each city using the feature maps following the second ConvLSTM
block (Figure~\ref{fig:gradcam_four_city}). The composites showed non-uniform
positive activation within all four city grids. Higher relative activation was
located broadly over the southern and western parts of the Bengaluru grid, the
central part of the Delhi grid, the central part of the Kolkata grid and the
western to south-western part of the Mumbai grid.

Each event-level Grad-CAM map was normalised independently before averaging,
and only positive gradient-weighted activations were retained. The composites
therefore show relative localisation within each selected model. They cannot be
compared as absolute contribution magnitudes across cities and do not identify
causal rainfall mechanisms or sub-grid atmospheric processes.

Event-to-event variability in the normalised ConvLSTM Grad-CAM fields is
presented in Supplementary Figure~S5.

\FloatBarrier

\clearpage
\begin{landscape}

\begin{table}[H]
    \centering
    \footnotesize
    \setlength{\tabcolsep}{4pt}
    \renewcommand{\arraystretch}{1.2}

    \caption{Summary of post-hoc XAI results for the validation-selected
    neural models.}
    \label{tab:xai_summary}

    \begin{tabularx}{\linewidth}{
        >{\raggedright\arraybackslash}p{1.8cm}
        >{\raggedright\arraybackslash}p{2.2cm}
        >{\raggedright\arraybackslash}X
        >{\centering\arraybackslash}p{2.4cm}
        >{\centering\arraybackslash}p{1.8cm}
        >{\centering\arraybackslash}p{2.4cm}
        >{\centering\arraybackslash}p{2.8cm}
    }

        \toprule

        \textbf{City}
        &
        \textbf{Model}
        &
        \textbf{Top predictor group}
        &
        \textbf{Group $\Delta$RMSE}
        &
        \textbf{Top input day}
        &
        \textbf{Lag $\Delta$RMSE}
        &
        \textbf{Maximum cell $\Delta$RMSE}
        \\

        \midrule

        Bengaluru
        & FC-LSTM
        & Wind
        & $0.63\pm0.07$
        & $t$
        & 1.50
        & 0.20 \\

        Bengaluru
        & ConvLSTM
        & Wind
        & $0.50\pm0.08$
        & $t$
        & 1.24
        & 0.07 \\

        \addlinespace

        Delhi
        & FC-LSTM
        & Past rainfall
        & $0.79\pm0.07$
        & $t$
        & 2.07
        & 0.39 \\

        Delhi
        & ConvLSTM
        & Past rainfall
        & $0.53\pm0.07$
        & $t$
        & 1.34
        & 0.14 \\

        \addlinespace

        Kolkata
        & FC-LSTM
        & Surface pressure
        & $0.57\pm0.12$
        & $t$
        & 1.96
        & 1.18 \\

        Kolkata
        & ConvLSTM
        & Wind
        & $0.45\pm0.27$
        & $t$
        & 1.37
        & 0.66 \\

        \addlinespace

        Mumbai
        & FC-LSTM
        & Past rainfall
        & $3.60\pm0.44$
        & $t$
        & 7.71
        & 2.46 \\

        Mumbai
        & ConvLSTM
        & Past rainfall
        & $5.79\pm0.43$
        & $t$
        & 8.82
        & 2.86 \\

        \bottomrule
    \end{tabularx}

    \begin{minipage}{0.94\linewidth}
        \vspace{3pt}
        \footnotesize
        \textit{Note:}
        All $\Delta$RMSE values are changes in domain-mean rainfall RMSE and
        are expressed in mm\,day$^{-1}$. Predictor-group values are means
        $\pm$ standard deviations across five permutation repetitions.
        Temporal- and spatial-occlusion values describe one
        validation-selected run per architecture and city. Test performance
        was not used to select these runs. Absolute changes are intended for
        within-city interpretation and should not be ranked as normalised
        sensitivity measures across cities.
    \end{minipage}
\end{table}

\end{landscape}
\clearpage

\section{Discussion}
\label{sec:discussion}

The comparison suggests that the relative performance of the forecasting
architectures depended partly on the spatial and temporal information retained
in the daily IMDAA fields. ConvLSTM did not consistently reduce prediction
error relative to the selected na\"ive references or the simpler fitted
models, and its relative performance differed across domain-mean rainfall,
complete-field and spatial-anomaly measures. This pattern is consistent with
the substantial difference between the small daily IMDAA domains used here and
the dense, high-frequency radar sequences for which convolutional recurrence
was originally developed \citep{Shi2015,Shi2017}. Arranging predictors as
spatial fields was therefore not, by itself, sufficient reason to prefer
ConvLSTM in this setting.

The choice of na\"ive reference materially affected the interpretation of
forecast skill. For domain-mean rainfall, grid climatology was the lower-RMSE
reference in Bengaluru and Kolkata, whereas persistence was the lower-RMSE
reference in Delhi and Mumbai. For spatial-anomaly RMSE, grid climatology was
the lower-RMSE reference in all four cities. Models that appeared to produce
substantial spatial improvements relative to persistence often produced only
small improvements relative to climatology. In Bengaluru, for example,
FC-LSTM achieved spatial-anomaly skill of 0.269 relative to persistence but
only 0.020 relative to climatology. In Delhi, the fitted models reduced
spatial-anomaly RMSE by approximately 28\% relative to persistence but by only
about 1--1.4\% relative to climatology. Reporting persistence-relative skill
alone would therefore have overstated the improvement over the stronger
seasonal reference \citep{Murphy1992}.

The three principal error measures also need to be interpreted jointly.
Complete-field RMSE combines errors in rainfall amount and within-field
variation, whereas domain-mean rainfall RMSE and spatial-anomaly RMSE separate
these components. Removing the daily domain mean isolates spatial variation,
but a low spatial-anomaly RMSE does not necessarily establish that a model has
reproduced the observed pattern with the correct amplitude. A smooth prediction
can obtain a favourable pointwise error when spatial contrasts are weak or
slightly displaced. This is consistent with precipitation-verification research
showing that pointwise scores can reward smoothing and penalise small spatial
displacements \citep{RobertsLean2008,Ebert2008}. Future work could complement
the present measures with spatial-anomaly correlation and
predicted-to-observed spatial-variance ratios, but the current results already
show why rainfall amount and spatial structure should not be represented by a
single score.

The architecture comparison indicates that local convolutional recurrence was
not consistently necessary at the scale of the evaluated fields. The inputs
had $0.12^{\circ}$ grid-point spacing, and the city domains contained only
9--42 cells. Sub-daily evolution was aggregated into one daily field, while
FC-LSTM processed the complete city field at every time step. In Kolkata, a
$3\times3$ convolutional kernel covered the full domain around the central grid
cell and a substantial proportion of the field at boundary cells, reducing the
practical distinction between local and domain-scale processing. FC-LSTM
produced the numerically lowest domain-mean rainfall and complete-field RMSE in
three cities, while ExtraTrees or CNN-only produced the lowest values for some
metrics in Delhi and Kolkata. These patterns indicate that recurrence without
local convolution, convolution without recurrence and non-neural full-field
mapping could represent much of the predictive information retained in the
daily inputs. They also agree with evidence that recurrent-model rankings vary
across locations and predictor formulations rather than following a stable
architecture hierarchy \citep{Panda2024}.

Mumbai provided the clearest numerical indication that convolutional recurrence
may be useful when recent spatial patterns remain informative. Its median
lag-1 spatial-anomaly correlation was 0.392, compared with values below 0.10 in
the other cities. Adding rainfall history reduced both domain-mean rainfall
RMSE and spatial-anomaly RMSE for all three neural architectures, and past
rainfall was the predictor group producing the largest permutation response
within both selected recurrent models. Temporal occlusion also showed that the
selected Mumbai models retained sensitivity beyond the latest input day.
ConvLSTM nevertheless improved on FC-LSTM by only about
0.13 mm\,day$^{-1}$ in spatial-anomaly RMSE, and that difference was not
subjected to paired uncertainty testing. The results are therefore consistent
with a role for short-term spatial continuity, but they do not establish it as
the cause of ConvLSTM's numerical ranking.

Other city-level differences prevent a stronger interpretation of the Mumbai
result. Mumbai differed in rainfall variability, domain shape, grid dimensions,
the share of variance associated with spatial anomalies and the change between
training- and test-period rainfall distributions. The comparison includes only
four cities and cannot isolate the contribution of spatial continuity from
these other characteristics. The result should therefore be treated as a
hypothesis for testing across additional regions, domain sizes and temporal
resolutions rather than as an established criterion for choosing ConvLSTM.

The rainfall-history comparison also shows that useful lagged information does
not necessarily require recurrent processing. Rainfall history made little
difference to the Bengaluru neural models, produced modest and
architecture-specific changes in Delhi, and improved CNN-only more than either
recurrent architecture in Kolkata. Because CNN-only received the seven previous
fields as stacked channels, it could use lagged rainfall without maintaining a
recurrent state. The contribution of rainfall history must therefore be
separated from the contribution of recurrence. These comparisons concern the
implemented configurations rather than entire model families: one
prespecified configuration was evaluated for each model, and the architectures
differed in structural assumptions and trainable parameter counts. Additional
tuning could change individual rankings, but it would not remove the need to
compare each fitted configuration with climatology, persistence and simpler
alternatives.

The most consistent weakness concerned high-rainfall days. All neural
architectures showed substantial negative bias and low detection rates above
the city-specific training-period $P_{90}$ threshold. Persistence achieved the
highest probability of detection and critical success index in every city,
while neural-model frequency bias was generally well below one. Low
false-alarm ratios for some neural models reflected the small number of
positive forecasts rather than reliable event detection. In Mumbai, for
example, FC-LSTM and persistence had similar high-rainfall RMSE, but FC-LSTM
had a much larger negative bias. Average error and high-rainfall reliability
were therefore distinct properties of model performance.

Several features of the experiment may have contributed to the upper-tail
underestimation. The logarithmic target transformation compressed differences
among larger rainfall values, and the unweighted MSE objective did not
explicitly reward correct prediction of threshold exceedances. High-rainfall
days also formed a minority of the training cases. These features may have
favoured conservative predictions, although the analysis does not isolate their
individual effects. The result is consistent with radar-nowcasting studies in
which deterministic models produced increasingly smooth fields and weaker
performance at higher rainfall intensities \citep{Ayzel2020,Ravuri2021}.
The test period was also wetter than the training period in Bengaluru, Delhi
and Mumbai, whereas Kolkata changed relatively little. This distributional
change may have compounded the underestimation but cannot be identified as its
cause. Intensity-sensitive, quantile and probabilistic objectives are relevant
extensions, but they should continue to be evaluated using both rainfall
magnitude and threshold-detection measures.

The post-hoc XAI results help explain fitted-model behaviour but do not alter
the benchmark conclusions. In Delhi, past rainfall produced the largest
grouped-permutation response within both selected recurrent models, yet neither
model improved on persistence for domain-mean rainfall. The networks depended
on previous rainfall but did not use it more effectively than direct
continuation. A large permutation response therefore indicates model reliance,
not forecast superiority. The grouped scores also concern unequal bundles of
information: the wind group contained zonal wind, meridional wind and directly
supplied wind speed, whereas the other meteorological groups contained one
channel. Wind-group responses should consequently be interpreted as reliance
on the complete group rather than as importance per variable. More generally,
correlated predictors can share or substitute information, and permutation
changes their observed relationships \citep{Strobl2008,Gregorutti2015,
Gregorutti2017}.

Within the validation-selected runs, temporal occlusion indicated that the
latest input day contributed most strongly to prediction error when removed.
Some additional sensitivity to recent lags remained in Kolkata and Mumbai.
This does not establish that older input days were unnecessary, because
one-at-a-time occlusion leaves correlated and potentially substitutable inputs
unchanged. Determining the required sequence length would require retraining
models with shorter input windows. The training-median replacements used for
temporal and spatial occlusion avoided arbitrary zero fields but did not
guarantee that the perturbed sequences remained within the joint distribution
of the training data.

Spatial-cell occlusion and Grad-CAM measured different aspects of model
behaviour. Occlusion recorded changes in prediction error after information at
a grid cell was replaced, whereas Grad-CAM localised positive activation
associated with the selected ConvLSTM output. Grad-CAM maps were independently
normalised by event and retained only positive gradient-weighted activation.
Their magnitudes therefore cannot be compared across cities, and they do not
identify causal rainfall processes, physically important locations or
monitoring priorities. The post-hoc XAI analyses also described only one
validation-selected run per architecture and city. Variation across
permutation repetitions is not uncertainty across independently trained
networks, and no model-randomisation or formal faithfulness test was performed.
The XAI results should therefore be interpreted as model-specific information
about fitted input--output relationships \citep{Adebayo2018,Ghorbani2019,
Ismail2020,Bommer2024,OLoughlin2025}.

The scope of the findings is limited to one-day-ahead deterministic hindcasting
during June--September for four city-centred IMDAA domains containing 9--42
grid cells. The predictors and target came from the same reanalysis and
data-assimilation system, so the reported errors measure prediction of IMDAA
precipitation rather than agreement with independent rain-gauge or radar
observations \citep{Rani2021}. The city grids differed in size and shape, and
the $0.12^{\circ}$ grid-point spacing does not support street- or ward-level
interpretation. The small domains may also omit approaching rainfall systems
outside their boundaries. Larger surrounding fields and sub-daily inputs could
provide convolutional models with more information about movement and
development. Only one prespecified configuration was evaluated for each model
family, and the post-hoc XAI analyses described one validation-selected run per
architecture and city.

Formal uncertainty in paired differences between forecasting methods was not
estimated. The reported neural-model standard deviations quantify sensitivity
to random initialisation, not uncertainty in the difference between two models
on the same test days. Close rankings---including the Mumbai
ConvLSTM--FC-LSTM spatial-anomaly difference and the near ties in Bengaluru,
Delhi and Kolkata---should therefore remain described as numerical rather than
as established superiority.

The findings support benchmark-led architecture selection for daily
rainfall-field prediction. A gridded input does not by itself justify
convolutional recurrence. In these small IMDAA domains, FC-LSTM, ExtraTrees,
CNN-only and the na\"ive references were frequently competitive with or better
than ConvLSTM, depending on the city and evaluation metric. Mumbai provided the
strongest numerical indication that ConvLSTM may benefit from more persistent
recent spatial patterns, but that association requires testing across larger
domains, additional regions and finer temporal resolutions. Model selection
should therefore begin with the predictive structure retained in the data and
with performance against credible na\"ive and simpler fitted alternatives,
rather than with an assumed hierarchy of model complexity.

\section{Conclusion}
\label{sec:conclusion}

This study tested whether convolutional recurrence reduces prediction error in one-day-ahead
daily IMDAA rainfall-field hindcasting over four small city-centred domains.
The results do not support a general ConvLSTM advantage. Model rankings varied
with the city, the forecast property and the reference forecast. FC-LSTM
produced the numerically lowest domain-mean RMSE in Bengaluru, Kolkata and
Mumbai, whereas persistence remained strongest in Delhi. ConvLSTM produced
the numerically lowest spatial-anomaly RMSE only in Mumbai, and its difference
from FC-LSTM was small and was not subjected to paired uncertainty testing.

The comparison against both persistence and grid climatology was essential.
Several models showed substantial positive spatial RMSE skill relative to
persistence, but only small improvements over climatology. Separating
rainfall amount from spatial anomalies similarly prevented a favourable score
for one property from being treated as general forecast superiority.

Mumbai provided the strongest indication that convolutional recurrence may
improve spatial prediction when successive daily fields retain local spatial
information. It had
the highest lag-1 spatial-anomaly continuity, rainfall history improved all
three neural architectures, and ConvLSTM achieved its strongest relative
spatial result there. This cross-city association is suggestive rather than
causal and requires testing across more regions, domain sizes and temporal
resolutions.

All neural models remained weak on high-rainfall days. They underestimated
rainfall above the city-specific training $P_{90}$ threshold and forecast too few
threshold exceedances. This limitation may reflect the logarithmic target
transformation, MSE optimisation, the rarity of upper-tail observations and
the wetter test-period distributions in Bengaluru, Delhi and Mumbai. Future
work should compare intensity-sensitive and probabilistic objectives while
retaining separate magnitude and detection measures.

The findings apply to IMDAA precipitation over domains containing 9--42 grid
cells and do not establish accuracy against independent rain-gauge or radar
observations. Within this setting, arranging meteorological predictors as maps
was not sufficient reason to prefer ConvLSTM. Convolutional recurrence should
therefore be evaluated against appropriate reference and comparison models,
rather than assumed to be preferable whenever predictors are spatially
arranged.

\section*{Data availability}
The IMDAA regional reanalysis fields used in this study are available through
the NCMRWF Reanalysis Data Service
(\url{https://rds.ncmrwf.gov.in/datasets}). Raw IMDAA files are not
redistributed with this article. The analysis used June--September data for
1998--2020. The code used for sequence construction, model training,
performance evaluation, post-hoc explainable artificial intelligence analyses
and figure generation is publicly available at
\url{https://github.com/TANMAY123484/Multicity-rainfall-forecasting-xai}.
The processed city-specific inputs required for reproduction are available
from the corresponding author upon reasonable request.

\section*{Statements and declarations}

\subsection*{Funding}
The authors received no specific funding for this work.

\subsection*{Competing interests}
The authors have no relevant financial or non-financial interests to disclose.

\subsection*{Author contributions}
Tanmay Ghosh: Conceptualisation, methodology, software, validation, formal
analysis, investigation, data curation, visualisation, writing--original draft,
and project administration. Shaurabh Anand and Rakesh Gomaji Nannewar: Data
curation, validation, and writing--review and editing. Nithin Nagaraj:
Writing--review and editing and project administration. All authors read and
approved the final manuscript.

\bibliographystyle{apalike}
\bibliography{bibliography}

\end{document}